\documentclass{article}

\usepackage[final,main]{neurips_2026}

\usepackage[utf8]{inputenc}
\usepackage[T1]{fontenc}
\usepackage{hyperref}
\usepackage{url}
\usepackage{booktabs}
\usepackage{longtable}
\usepackage{amsmath,amssymb,amsthm,amsfonts,mathtools,bm}
\usepackage{graphicx}
\usepackage{xcolor}
\usepackage{colortbl}
\usepackage{capt-of}
\usepackage{tikz}
\usepackage{microtype}
\usepackage{algorithmic}
\usetikzlibrary{calc}

\newtheorem{theorem}{Theorem}
\newtheorem{proposition}{Proposition}
\newtheorem{corollary}{Corollary}

\newtheorem{remark}{Remark}

\newcommand{\TV}{\mathrm{TV}}
\newcommand{\Var}{\mathrm{Var}}
\newcommand{\Cov}{\mathrm{Cov}}
\newcommand{\tr}{\mathrm{tr}}
\newcommand{\E}{\mathbb{E}}
\newcommand{\Prob}{\mathbb{P}}
\newcommand{\Law}{\mathcal{L}}
\newcommand{\mstd}[2]{#1 {\scriptsize$\pm$ #2}}
\definecolor{taxorange}{HTML}{F39A2E}
\definecolor{taxblue}{HTML}{113F83}
\definecolor{taxlink}{HTML}{9A6803}
\definecolor{taxhighlight}{HTML}{F39A2E}
\hypersetup{colorlinks=true, linkcolor=taxlink, citecolor=taxblue, urlcolor=taxlink}
\newcommand{\strongnum}[1]{\begingroup\setlength{\fboxsep}{1.2pt}\fcolorbox{taxhighlight}{taxhighlight!18}{\strut #1}\endgroup}

\newenvironment{paperblock}{\par\vspace{0.25em}\noindent\begingroup\centering\ignorespaces}{\unskip\par\endgroup\vspace{0.25em}}

\title{Serialization Tax in Shared-Latent Exchangeable Decisions}
\hypersetup{
  pdftitle={Serialization Tax in Shared-Latent Exchangeable Decisions},
  pdfauthor={Siming Zhang, Zhehui Shen, Shijie Chen, Xinle Gu, Yansen Yu, Hang Yu}
}

\author{
  Siming Zhang \quad Zhehui Shen \quad Shijie Chen \\
  \textbf{Xinle Gu \quad Yansen Yu \quad Hang Yu}\thanks{Corresponding author: \texttt{yuhang@shu.edu.cn}.} \\
  Shanghai University
}

\begin{document}

\maketitle
\raggedbottom
\setlength{\abovecaptionskip}{3pt}
\setlength{\belowcaptionskip}{2pt}

\begin{abstract}
Accurate item probabilities can produce inaccurate group-level decisions. A system may score each item separately, yet act on whether enough hidden items succeed: show a slate, defer a panel decision, or inspect a batch. When items share an uncertain latent state, independent marginalization preserves means but removes dependence from the aggregate posterior. We call this error the serialization tax. For binary items sharing one success rate, classical convex-order comparisons give a more dispersed exact count law than any independent completion with the same total mean. We quantify variance and total-variation gaps, derive threshold separation under margin conditions, and connect posterior approximation to decision regret. Separate results cover non-binary outcomes, heterogeneous items, and sufficient posterior resolution. TaxScore identifies large hidden blocks with unresolved uncertainty near a decision boundary for routing to richer posterior heads. On pre-specified, hidden-target-free MovieLens 1M/10M audits, existing shared-latent heads improve high-score negative log-likelihood (NLL)/utility from $0.673/0.696$ to $0.629/0.723$ and from $0.670/0.691$ to $0.639/0.715$. Routing gives $0.006$--$0.012$ full-population utility gains, with matched controls identifying their sources. Our code is available at \url{https://github.com/SiminChang/Serialization-Tax-in-Shared-Latent-Exchangeable-Decisions}.
\end{abstract}

\section{Introduction}
Many machine-learning pipelines predict one score per item, as in recommendation and implicit-feedback modeling \citep{covington2016youtube,hu2008implicit,rendle2009bpr}. A downstream action can depend on the group: whether a slate contains enough relevant movies, a measurement panel warrants deferral, or a batch warrants inspection. Such decisions require the probability of an aggregate event. If hidden items share an uncertain user or batch state, their posterior predictions remain dependent after that state is marginalized. Replacing the joint predictive distribution by the product of its marginals preserves every item mean but removes this dependence. The resulting aggregate probability can therefore be inaccurate even when each item probability is calibrated.

Posterior predictive correlations and joint prediction benchmarks already show why uncertainty beyond marginals matters \citep{wang2021beyondmarginal,osband2022neuraltestbed}. We study the specific approximation that creates the discrepancy: averaging a shared latent state separately for each item before computing a group outcome. Retaining that state through aggregation yields a mixture of conditional count laws with different tail behavior. Classical stochastic-order results describe part of this change \citep{cousin2008comparison,hoeffding1956}. We connect this ordering to posterior gaps, decision diagnostics, and an audit of the decision value and measured computational cost of shared posterior inference.

\begin{paperblock}
\begingroup
\scriptsize
\resizebox{0.88\linewidth}{!}{%
\begin{tikzpicture}[
  font=\scriptsize,
  box/.style={draw=black!35, rounded corners=2pt, fill=taxorange!6, minimum width=5.48cm, minimum height=4.38cm},
  item/.style={draw=taxblue!65, circle, fill=taxblue!8, minimum size=0.62cm, inner sep=1pt},
  latent/.style={draw=taxorange!85!black, circle, fill=taxorange!22, minimum size=0.74cm, inner sep=1pt},
  arrow/.style={->, thick, taxblue!70},
  call/.style={draw=black!25, rounded corners=2pt, fill=taxblue!5, align=center, inner sep=3pt},
  pill/.style={draw=taxorange!80!black, rounded corners=2pt, fill=taxorange!12, align=center, inner sep=3.5pt}
]
\node[call, text width=4.65cm, fill=taxblue!6] (same) at (0,2.62)
  {\textbf{itemwise marginals identical}\\$\Prob(X_1=1)=\Prob(X_2=1)=.50$};
\node[box] (shared) at (-3.15,0) {};
\node[box] (indep) at (3.15,0) {};
\node[font=\bfseries] at (-3.15,1.72) {Exact shared-latent posterior};
\node[font=\bfseries] at (3.15,1.72) {Independent marginal completion};
\node[latent] (q) at (-3.15,0.95) {$Q$};
\node[item] (sx1) at (-4.35,0.16) {$X_1$};
\node[item] (sx2) at (-1.95,0.16) {$X_2$};
\draw[arrow] (q) -- (sx1);
\draw[arrow] (q) -- (sx2);
\node[align=center] at (-3.15,-0.72) {one posterior draw controls both items\\$Q=0$ or $Q=1$ with probability $.50$};
\node[item] (iz1) at (2.20,0.92) {$Z_1$};
\node[item] (iz2) at (4.10,0.92) {$Z_2$};
\node[item] (ix1) at (2.20,-0.34) {$X_1$};
\node[item] (ix2) at (4.10,-0.34) {$X_2$};
\draw[arrow] (iz1.south) -- (ix1.north);
\draw[arrow] (iz2.south) -- (ix2.north);
\node[align=center] at (3.15,-1.05) {two separate $.50$ coins\\shared uncertainty already averaged out};
\node[align=center] at (-3.15,-1.50) {$\Prob(R_2=2)=.50$};
\draw[fill=taxblue!72, draw=taxblue!72] (-4.45,-1.82) rectangle (-3.15,-1.66);
\draw[draw=black!35] (-4.45,-1.82) rectangle (-1.85,-1.66);
\node[align=center] at (3.15,-1.50) {$\Prob(\widehat R_2=2)=.25$};
\draw[fill=taxblue!72, draw=taxblue!72] (1.85,-1.82) rectangle (2.50,-1.66);
\draw[draw=black!35] (1.85,-1.82) rectangle (4.45,-1.66);
\node[pill, text width=2.68cm] at (-3.55,-3.02) {\textbf{same marginals}\\itemwise calibration unchanged};
\node[pill, text width=2.68cm] at (0,-3.02) {\textbf{tail drops}\\$.50\to .25$};
\node[pill, text width=2.68cm] at (3.55,-3.02) {\textbf{diagnostic}\\$\TV\ge v(O)=1/4$\\TaxScore high};
\end{tikzpicture}}

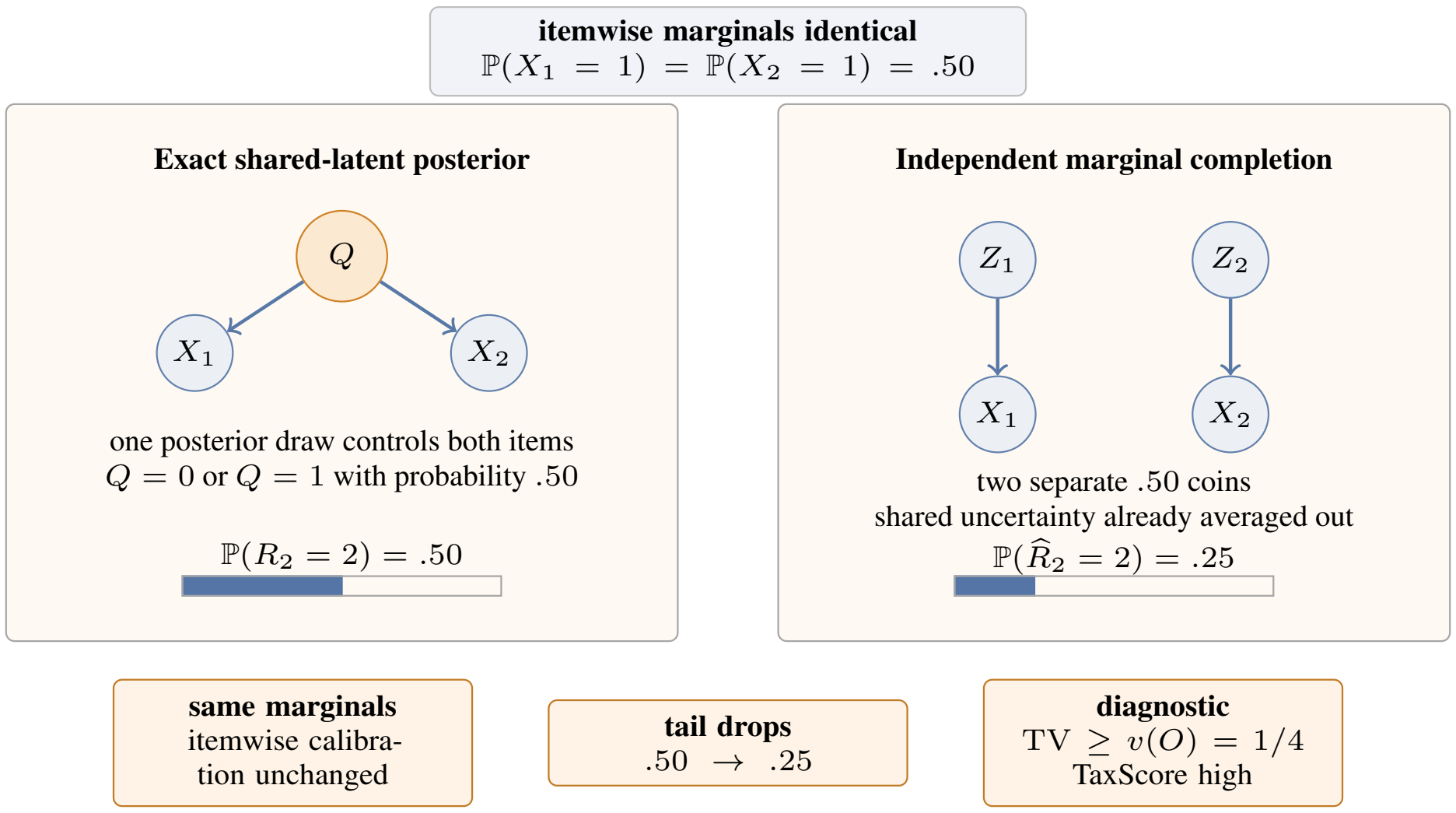
\captionof{figure}{Same marginals, different aggregate posterior ($n=2$). One shared latent draw gives identical outcomes, whereas independent completion uses two separate coins. Both have item probability $.50$, but the probability of two successes falls from $.50$ to $.25$. The count-law total-variation (TV) distance is $.50$, exceeding the lower bound $v(O)=1/4$ in Theorem~\ref{thm:universal}. Locally identical predictions thus give different probabilities for the event used by the decision.}
\label{fig:intro-joint}
\endgroup
\end{paperblock}

Figure~\ref{fig:intro-joint} illustrates the resulting aggregate-posterior error, which we call the \emph{serialization tax}. Given observations $O$, hidden-item covariates $W_{1:n}$, and $n=m-t$ hidden items, the decision may require $\Prob(R_n\ge\lceil cn\rceil\mid O,W)$, where $R_n=\sum_{i=t+1}^m X_i$. A model can estimate this probability directly or obtain it from shared posterior samples, using either a sequence or set backbone. The analysis concerns the probabilistic information passed to the decision. The MovieLens study tests a pre-specified regime with ambiguous prefixes and near-threshold slates and reports both selected-slice and full-population effects under a fixed decision rule and a declared inference budget.

Our contributions are threefold:
\begin{itemize}
\setlength{\itemsep}{0pt}
\setlength{\topsep}{2pt}
\item \textbf{To our knowledge, this is the first unified formulation and analysis of this averaging-before-aggregation failure for shared-latent aggregate tail decisions.} We quantify lost count uncertainty through exact variance identities, TV lower bounds, and equality conditions, building on classical convex-order comparisons. Separate empirical-measure and heterogeneous-item results identify the appropriate comparison for each decision object.
\item \textbf{Decision conditions and sufficient posterior resolution.} Threshold-margin bounds motivate TaxScore. Finite-probe lower bounds and grid-resolution guarantees specify the information needed for tail prediction. A bounded-regret result connects posterior accuracy to action value while making the role of the utility and decision boundary explicit.
\item \textbf{A reproducible routing audit with attribution and cost controls.} Controlled bags verify mechanism scaling, and MovieLens audits show selected-slice NLL/utility improvements and $0.006$--$0.012$ full-population gains. Matched-backbone, fixed-ensemble, equal-budget, and latency controls separate probability quality, decision value, and computational cost.
\end{itemize}

\section{Background and Related Work}
\paragraph{Exchangeability and stochastic orders.}
Exchangeability and de Finetti theory represent an infinitely exchangeable sequence through a random probability measure \citep{definetti1937,hewitt1955,kallenberg2005}. The homogeneous mixture-versus-binomial ordering is a classical dependence comparison \citep[Theorem~2.11]{cousin2008comparison}. Hoeffding's comparison covers independent Bernoulli sums with the same total mean \citep[Theorem~3]{hoeffding1956}. Stochastic orders and majorization provide the broader context \citep{shaked2007stochastic,marshall2011majorization,barbour1992poisson,hong2013poissonbinomial}. We use this lineage explicitly, then derive conditional posterior identities, threshold bounds, and resolution conditions that connect the comparison to a decision audit with an explicit action, cost model, and inference budget.

\paragraph{Joint predictive uncertainty.}
\citet{wang2021beyondmarginal} evaluate posterior predictive correlations for regression and active learning, while the Neural Testbed evaluates joint predictions across agents and environments \citep{osband2022neuraltestbed}. Our focus is complementary: identify the marginalization operation that removes dependence, quantify its consequences for aggregate tails, and test whether a decision-time diagnostic concentrates the observed decision value of joint prediction.

\paragraph{Architectures, inference, and tail decisions.}
Deep Sets, Janossy pooling, Set Transformers, flows, neural processes, dropout, and ensembles model exchangeable data or predictive uncertainty \citep{zaheer2017deepsets,murphy2019janossy,lee2019set,rasul2019setflow,zwartsenberg2022cpif,garnelo2018cnp,garnelo2018np,kim2019anp,gal2016dropout,lakshminarayanan2017simple}. Each backbone still specifies an output object: marginals, a count distribution, or joint samples. Slate bandits and ranking \citep{dimakopoulou2019marginal,aouali2021combining}, multiple-instance learning \citep{dietterich1997solving,briggs2012rank}, and calibration \citep{pmlr-v80-hebert-johnson18a,kuleshov2018accurate,vovk2005alrw} motivate distinct objectives. Stop-loss, chance constraints, and conditional value-at-risk (CVaR) depend on aggregate tails \citep{charnes1959chance,rockafellar2000cvar}. Additive expected relevance uses only marginal means, so the utility determines which posterior information the action requires and where our dependence analysis applies.

\paragraph{Task objectives and changing populations.}
Beyond-accuracy recommendation studies jointly evaluate accuracy, diversity, and serendipity \citep{chen2024cognitive}, illustrating why an audit must declare its action and reward. Our experiments fix a tail event and evaluate its predict/defer payoff. Diversity-aware slate construction requires a separate objective-dependent analysis. Group concept-drift detection studies changes shared across data streams \citep{yu2023groupdrift}. It addresses temporal distribution change, while our analysis conditions on a given posterior. Together these perspectives motivate monitoring whether a routing score calibrated on one population continues to select valuable joint inference after changes in user composition, item availability, or the operating reward.

Our work connects a precise approximation, decision-specific error bounds, and an audit of value and computational cost. Together, these components turn classical dependence comparisons into a reproducible test of when retaining shared posterior uncertainty benefits the downstream action.

\section{Shared-Latent Counts and Independent Marginalization}
The analysis begins with the number of hidden successes after a shared latent state is averaged separately for each item. Binary counts admit a convex-order comparison. Non-binary observations and heterogeneous covariates require separate mathematical statements, each describing the same approximation through the aggregate object and loss function used by the downstream decision.

Let $O$ reveal $t$ coordinates of a binary sequence $X_{1:m}$, leaving $n=m-t$ hidden items. Assume one uncertain success rate is drawn from $Q\mid O\sim\nu_O$, then shared by all hidden items: $X_{t+1},\ldots,X_m\mid Q,O\overset{i.i.d.}{\sim}{\rm Bernoulli}(Q)$. This hierarchical mixture is the binary form of de Finetti's representation. The shared-rate assumption is stronger than finite permutation invariance. Write $R_n=\sum_{i=t+1}^m X_i$, $\bar q=\E[Q\mid O]$, and $v(O)=\Var(Q\mid O)$. We use
\[
\TV(P,Q)=\sup_A |P(A)-Q(A)|=\frac12\|P-Q\|_1=\sup_{0\le g\le 1}|\E_Pg-\E_Qg|.
\]
A total-variation (TV) gap is therefore a worst-case bounded test-function gap.
A \emph{mean-preserving independent completion} replaces the hidden items by independent $\widetilde X_i$ with means $p_i(O)$ satisfying $\sum_i p_i(O)=n\bar q$, and writes $\widehat R_n=\sum_i\widetilde X_i$.

The ordering below combines \citet[Theorem~2.11]{cousin2008comparison} with \citet[Theorem~3]{hoeffding1956}. The variance identity, TV bound, and equality cases quantify this conditional comparison.
\begin{theorem}[Convex-order comparison and posterior gaps]
\label{thm:universal}
For every observation $O$ and every discrete convex function $\phi:\{0,\ldots,n\}\to\mathbb R$,
\[
\E[\phi(R_n)\mid O]\ge \E[\phi(\widehat R_n)\mid O].
\]
Moreover, for all $n\ge1$,
\begin{align}
\Var(R_n\mid O)-\Var(\widehat R_n\mid O)
&=n(n-1)v(O)+\sum_i(p_i(O)-\bar q)^2. \label{eq:var-gap-main}
\end{align}
If $n\ge2$, then
\begin{align}
\TV(\Law(R_n\mid O),\Law(\widehat R_n\mid O))&\ge v(O). \label{eq:tv-gap-main}
\end{align}
For $n\ge2$, equality of the two posterior count laws holds iff $v(O)=0$ and $p_i(O)=\bar q$ for every hidden item. For $n=1$, both count laws are ${\rm Bernoulli}(\bar q)$, so the count-law TV gap vanishes even when $v(O)>0$.
\end{theorem}

\paragraph{Takeaway.}
Independent completion can preserve the aggregate predictive mean, and even local calibration, while understating count uncertainty. Jensen gives the homogeneous mixture comparison and Hoeffding handles heterogeneous independent means. Convex losses, including variance and stop-loss, inherit this ordering. Threshold indicators require separate margin conditions because they are nonconvex. Decision regret also depends on the utility and whether the posterior change moves the optimal action. Appendix~\ref{app:deferred} develops these consequences and perturbation bounds that quantify departures from the exact shared-latent assumptions in finite populations.

Rating bins, continuous measurements, and multi-class panels require a comparison beyond binary counts. The appropriate non-binary statement is a Laplace-functional order for the hidden empirical measure. This choice preserves a valid distributional comparison in settings where unrestricted multivariate convex order fails.

\begin{theorem}[Full empirical-measure law]
\label{thm:partition}
Let $(X_i)$ be infinitely exchangeable on a standard Borel space with directing random measure $G$, hidden empirical measure $M_n=\sum_{i=t+1}^m\delta_{X_i}$, and posterior mean measure $\bar G_O=\E[G\mid O]$. Let $\widehat M_n=\sum_i\delta_{\widetilde X_i}$ be any independent completion whose average marginal is $\bar G_O$. Then for every bounded measurable $h$,
\begin{align}
\E\!\left[e^{\int h\,dM_n}\middle|O\right]
&= \E\!\left[\left(\int e^{h(x)}dG(x)\right)^n\middle|O\right] \label{eq:partition-exact}\\
&\ge \left(\int e^{h(x)}d\bar G_O(x)\right)^n \label{eq:partition-iid}\\
&\ge \E\!\left[e^{\int h\,d\widehat M_n}\middle|O\right]. \label{eq:partition-approx}
\end{align}
Laplace functionals determine the law of a finite random measure, so this is the full non-binary object-level extension. A literal unrestricted multivariate convex-order theorem is false in dimension $d\ge3$. Appendix Proposition~\ref{prop:nocx} gives the counterexample. In applications, $h$ or a finite partition is chosen to match the decision, such as low/mid/high rating bins or measurement threshold bands.
\end{theorem}

For an application, the audit chooses $h$, a partition, or a statistic that matches the downstream tail event. Rating bins, count truncations, and measurement threshold bands provide concrete choices. Fixing this object connects the empirical-measure comparison to the posterior quantity used by the action.

\section{Heterogeneous Slates and Decisions}
Real slates are heterogeneous: movies have genres, products have covariates, and items can respond differently to the same latent state. The following identity quantifies the covariance removed by independent marginalization, conditional on both the observed prefix and visible item features.

\begin{proposition}[Covariate-conditional shared-latent tax]
\label{prop:covlatent}
Condition on $O$ and hidden item covariates $W_{1:n}$. Let $\Theta\mid O,W_{1:n}$ be a shared latent state and let $Y_i\mid\Theta,O,W\sim{\rm Bernoulli}(q_i(\Theta))$ independently. For $R=\sum_iY_i$ and the independent marginal completion $\widehat R=\sum_i\widehat Y_i$, $\widehat Y_i\sim{\rm Bernoulli}(\E[q_i(\Theta)\mid O,W])$,
\[
\Var(R\mid O,W)-\Var(\widehat R\mid O,W)=\sum_{i\ne j}\Cov(q_i(\Theta),q_j(\Theta)\mid O,W).
\]
If the posterior over $\Theta$ is associated \citep{esary1967association} and the $q_i$ are coordinatewise nondecreasing, all terms are nonnegative. More generally, suppose a posterior region $\mathcal H$ of mass $\alpha$ puts $\E[R\mid\Theta,O,W]$ at least $\Delta_{\rm count,\mathcal H}>0$ above threshold $a$, while $\E[\widehat R\mid O,W]$ is $\Delta_{\rm count,ind}>0$ below $a$. The exact tail exceeds the independent tail by at least
\[
\alpha(1-e^{-2\Delta_{\rm count,\mathcal H}^2/n})-e^{-2\Delta_{\rm count,ind}^2/n}.
\]
\end{proposition}

These margins are in \emph{count units}. A probability margin $\rho$ corresponds to $\Delta_{\rm count}=n\rho$, so $e^{-2\Delta_{\rm count}^2/n}=e^{-2n\rho^2}$. The identity retains shared covariance under heterogeneous item features. Its sign determines whether independent marginalization reduces or increases aggregate variance.

\begin{proposition}[Signed covariance audit]
\label{prop:signedcov}
With $C_{ij}=\Cov(q_i(\Theta),q_j(\Theta)\mid O,W)$ and $S_+=\sum_{i\ne j}C_{ij,+}$, $S_-=\sum_{i\ne j}C_{ij,-}$, the variance tax equals $S_+-S_-$. If $S_-\le\eta S_+$, it is at least $(1-\eta)S_+$. If negative covariance is comparable to positive covariance, the diagnostic must retain its sign to assess the net aggregate-variance change.
\end{proposition}

To connect posterior accuracy to a decision, fix the observed information $(O,W)$ and let $Z$ denote the decision-relevant outcome, such as the hidden count or its threshold indicator.
\begin{theorem}[Bayes value transfer]
\label{thm:decision}
Let $P=\Law(Z\mid O,W)$ be the target conditional law and $\widehat P$ the predictive law used by the decision. For a common action set and utility $u(a,z)\in[0,1]$, assume Bayes maximizers $a_P\in\arg\max_a\E_Pu(a,Z)$ and $a_{\widehat P}\in\arg\max_a\E_{\widehat P}u(a,Z)$ exist. Then, conditionally on $(O,W)$,
\[
0\le \E_Pu(a_P,Z)-\E_Pu(a_{\widehat P},Z)\le 2\TV(P,\widehat P).
\]
\end{theorem}
This is an \emph{upper bound}: posterior closeness controls regret, while large TV can leave the optimal action unchanged. Establishing positive regret requires an additional action-crossing argument. For binary predict/defer decisions, Appendix Proposition~\ref{prop:reveal} supplies such conditions and regret lower bounds. MovieLens measures realized payoff under validation-tuned thresholds, whose departure from Bayes optimality contributes a separate residual (Appendix~\ref{app:decision-scope}). Expected stop-loss and CVaR comparisons follow directly from convex order through their convex aggregate-risk representations.

\section{Diagnostics and Posterior Resolution}
But how can we identify the decisions that benefit most from shared posterior computation? A large tax needs three ingredients: a hidden block large enough to amplify dependence, unresolved shared-latent uncertainty, and a decision boundary close to the posterior mass. For a threshold $Y_n(c)=\mathbf1[R_n\ge\lceil cn\rceil]$, the gap can be constant even as $n$ grows. If $Q\mid O=(1-\alpha)\delta_{q_0}+\alpha\delta_{q_1}$ with $q_0<c<q_1$, then the exact threshold posterior tends to $\alpha$, while local i.i.d. completion tends to $\mathbf1[(1-\alpha)q_0+\alpha q_1>c]$.

\begin{proposition}[TaxScore diagnostic]
\label{prop:taxscore}
\label{prop:threshold}
In the two-point setting above, define the probability margins $\rho_-(O;c)=\min\{q_1-c,c-\bar q\}$ when $\bar q<c$ and $\rho_+(O;c)=\min\{\bar q-c,c-q_0\}$ when $\bar q>c$. With $\mathsf{TaxScore}_n=n\rho_-^2$ or $n\rho_+^2$ respectively, the finite-$n$ threshold-posterior gap is at least $\alpha-(1+\alpha)e^{-2\mathsf{TaxScore}_n}$ in the first case and $(1-\alpha)-2e^{-2\mathsf{TaxScore}_n}$ in the second. Thus the visible scale is $n v(O)$ when margins are proportional to $\sqrt{v(O)}$. The count margin $n\rho_\pm$ recovers Proposition~\ref{prop:covlatent}'s exponent under the same bounded-summand concentration inequality.
\end{proposition}
\begin{corollary}[Finite-sample threshold separation]\label{cor:thresholdrate}
The displayed TaxScore bounds are the Hoeffding finite-sample form \citep{hoeffding1963probability} of the constant asymptotic threshold separation. Appendix~\ref{app:proofs} states and proves both directions of the threshold separation for finite hidden blocks.
\end{corollary}
\begin{remark}[Effective ambiguity scale]\label{rem:nv-scale}
The diagnostic combines the hidden block size, unresolved latent ambiguity, and the margin to the particular tail event. Each component matters through its effect on the tail probability, and hence on the action selected under the stated utility.
\end{remark}

The theoretical score is specific to the two-point model and its crossing assumptions. An empirical score combining ambiguity, block size, and boundary distance is a \emph{surrogate} whose usefulness requires evaluation. Large variance can leave an action unchanged far from the boundary, while collapsed uncertainty can make marginals sufficient nearby. We test whether the score concentrates useful joint-prediction gains, with low-score groups as controls. A direct Transformer count head may already learn the relevant tail accurately. Applying the lower bounds below requires establishing that its output is determined by a fixed dictionary of posterior expectations across the compared instances.

Fixed finite summaries cannot uniformly recover threshold posteriors. A $\Psi$-probe compression is an output determined only by the vector of posterior expectations $(\int\psi_\ell\,d\nu)_\ell$. Let the dictionary include $\psi_0\equiv1$, and let a finite witness support $\mathcal X=\{x_1,\dots,x_J\}\subset[0,1]\setminus\{c\}$ have evaluation matrix $M_{\ell j}=\psi_\ell(x_j)$ and threshold vector $b_j=\mathbf1[x_j>c]$. If $b$ lies outside the row span of $M$, two posterior laws match all probes but have different probability mass above the threshold.

\begin{paperblock}
\begin{minipage}{0.985\linewidth}
\captionof{table}{Predictive representations and what the analysis establishes. Each guarantee applies to the stated predictive representation.}
\label{tab:audit}
\footnotesize
\setlength{\tabcolsep}{2pt}
\renewcommand{\arraystretch}{0.96}
\begin{tabular}{@{}p{0.22\linewidth}p{0.19\linewidth}p{0.55\linewidth}@{}}
\toprule
Representation & Formal support & Decision implication \\
\midrule
Independent marginals & Theorem~\ref{thm:universal}. Prop.~\ref{prop:covlatent} & Preserves local means but removes shared-latent covariance before aggregation, action pricing, and threshold evaluation. \\
Fixed probe summaries & Prop.~\ref{prop:probe}. Cor.~\ref{cor:minimax} & Threshold ambiguity persists under matched probe expectations when the stated row-span conditions hold. \\
TaxScore triage & Prop.~\ref{prop:taxscore} & Gives a two-point-model tail bound. Empirical scores require separate validation of routing gains. \\
Shared-state integration & Prop.~\ref{prop:resolution} & Resolving posterior mass near the boundary controls grid error. Learned latent heads are tested empirically. \\
Value audit & Theorem~\ref{thm:decision} & Posterior closeness bounds Bayes regret. Positive value requires an action change under the specified utility. \\
\bottomrule
\end{tabular}
\end{minipage}
\end{paperblock}

\begin{proposition}[Finite-probe lower bound]
\label{prop:probe}
Under the finite-support, constant-probe, and row-span conditions above, every $\Psi$-probe compression has asymptotic threshold-posterior error at least half the induced difference in posterior mass above $c$ on one of the two matched instances. A suitable bounded-utility decision problem also has a positive worst-case loss, with the utility chosen to distinguish the two matched posteriors.
\end{proposition}
\begin{corollary}[Minimax probe-compression tax]\label{cor:minimax}
Under the finite-support row-span condition, every $\Psi$-probe compression has strictly positive worst-case asymptotic threshold-posterior risk.
\end{corollary}

Escaping the lower bound requires resolution at the decision scale. Let $T_n(q)=\Prob({\rm Bin}(n,q)\ge\lceil cn\rceil)$ and let $\pi_h$ be a grid projection with $|\pi_h(q)-q|\le h$.
\begin{proposition}[Resolution needed to escape the threshold tax]
\label{prop:resolution}
For every $\gamma>h$,
\[
\left|\int T_n(q)d\nu(q)-\int T_n(\pi_h(q))d\nu(q)\right|
\le \nu\{|Q-c|\le\gamma\}+2e^{-2n(\gamma-h)^2}.
\]
Thus a dimension-growing grid or mixture head escapes when its mesh resolves the decision margin and little posterior mass lies in the unresolved band.
\end{proposition}
The restriction concerns the information retained by the representation. A finite-dimensional stochastic latent head can carry threshold-relevant posterior uncertainty. A large deterministic head restricted to fixed probes or independent marginals can still discard the mass needed to recover the tail probability.

\paragraph{Finite-probe scope.}
The result concerns summaries determined by a fixed dictionary of posterior expectations, such as finitely many moments. A learned count head or rollout state satisfies it only when that information restriction is established. Support-adaptive, dimension-growing, or stochastic representations can retain the missing threshold information, and a task-specific head can directly predict the required tail from the observed group under a task-specific supervised objective.

\paragraph{Computing the aggregate before averaging.}
Low-dimensional shared latents can be resolved by grid or mixture heads. Higher-dimensional latents can use stochastic global-latent posteriors such as neural-process or flow heads. In each case, one posterior draw, mixture component, or ensemble member is propagated across all hidden items before computing the tail probability. The same joint predictive computation applies to recurrent sequence and permutation-invariant set backbones.

\section{Experiments}
\label{sec:experiments}
\paragraph{Experimental setting.}
We organize the experiments around four linked tests: \emph{mechanism scaling}, \emph{diagnostic selection}, \emph{gain attribution}, and \emph{routing value}. Controlled synthetic/Covtype bags test scaling with ambiguity \citep{blackard1998covertype}. MovieLens 1M/10M test whether selected slates benefit from joint prediction, which controls explain the gain, and what compute routing requires \citep{harper2015movielens}. Natural multiple-instance-learning (MIL) tasks supply boundary cases \citep{dietterich1997solving,briggs2012rank}. Every MovieLens example reveals $k$ ratings and predicts whether at least $\lceil cn\rceil$ of $n$ hidden ratings are at least four stars. All heads share five seed-specific user splits and are evaluated on identical revealed prefixes, hidden slates, and tail-event labels.

\emph{Models and metrics.} Baselines include independent completion, gated recurrent unit (GRU) and Transformer count heads \citep{cho2014learning,vaswani2017attention}, rollout predictors \citep{NEURIPS2025_e31fc00e}, set heads \citep{zaheer2017deepsets,lee2019set}, and parameter-matched controls. Joint predictions use grid mixtures, an Attentive Neural Process (ANP) \citep{kim2019anp}, conditional flows \citep{rezende2015flows,rasul2019setflow,zwartsenberg2022cpif}, or two-tower ensemble mixtures \citep{covington2016youtube,lakshminarayanan2017simple}. Tail NLL/Brier measure probability quality. Utility measures action value. Full-population switch gain and latency measure deployment cost. Reliability and expected calibration error (ECE) supplement these metrics. MovieLens utility scores a correct confident decision as $1$, an error as $0$, and deferral as $1-\lambda$ with primary cost $\lambda=.30$. Validation users determine the confidence threshold. This normalized predict/defer payoff evaluates action value. Normalized discounted cumulative gain (NDCG) and Recall evaluate ranking relevance, which has a different action space and objective from the threshold triage task studied here.

\paragraph{Decision-time scoring.}
The predict/defer payoff gives a concrete interpretation of posterior refinement. At $\lambda=.30$, a Bayes decision defers when the larger event probability is below $.70$. For example, changing a predicted success probability from $.69$ to the correct $.72$ replaces deferral with a positive prediction, increasing its expected payoff from $.70$ to $.72$. A refinement from $.80$ to $.85$ leaves the positive action unchanged. These examples explain why improved tail NLL and improved realized utility are distinct outcomes and why the audit reports both on the same groups.

Controlled studies use the known data-generating posterior. MovieLens uses two distinct steps. Candidate construction accepts a slate with training-popularity mean $\bar p_{\rm prior}$ within $.08$ of $c$ and both revealed outcomes present, or minimizes $|\bar p_{\rm prior}-c|+.20|\bar y_{\rm obs}-1/2|$ over $80$ candidates. Genres enter the prediction heads, while the selector uses popularity and revealed outcomes. A fixed scalar-user grid posterior, updated only from these available signals, then supplies hidden-slate component means $\mu_g$ and weights $w_g$. The empirical routing score is
\[
S(O)=\frac{n\sum_g w_g(\mu_g-\bar\mu)^2}{(|\bar\mu-\lceil cn\rceil/n|+.03)^2},\qquad \bar\mu=\sum_g w_g\mu_g.
\]
The routing score estimates posterior ambiguity relative to the event boundary. It is distinct from candidate selection and from Proposition~\ref{prop:taxscore}'s two-point bound. Its inputs are training popularity, item identities, and revealed outcomes, excluding hidden targets and predictions from the compared neural heads. Descriptive bins and top-$p$ slices use unlabeled test-score ranks. The original switch transfers a numeric validation-score cutoff, while the equal-budget comparison selects an exact fraction within each split. Fractions and algorithms are fixed before evaluation, with the complete grid estimator, split-specific cutoffs, and allowed score inputs documented in Appendix~\ref{app:movie-score}.

\paragraph{Results.}
Controlled errors grow with hidden-block ambiguity, MovieLens gains concentrate on flagged slates, and matched controls separate probability improvements from utility changes. Routing then retains modest population gains at a reduced rich-head budget. These experiments evaluate the diagnostic in a deliberately selected high-risk regime and identify the conditions under which additional inference improves probability quality and changes a cost-sensitive action.

\paragraph{Law scaling.}
Synthetic and covtype bags instantiate the theorem-aligned regime. Appendix Figures~\ref{fig:scaling}--\ref{fig:oracle} show threshold-posterior, TV, and variance gaps growing with hidden size and latent posterior variance. On full covtype after revealing $k=3$ items, event-TV rises $0.070\to0.104$ and independent-completion NLL $0.627\to1.220$ as $n$ grows from $16$ to $96$. Stronger sequence heads reduce this error, while object-space prediction retains an advantage on the high-ambiguity prefixes. At $n=96$, IndepPB NLL rises from $.533$ to $1.945$ between low- and high-$v(O)$ prefixes, while ObjSetTF rises from $.329$ to $.728$. Figure~\ref{fig:covtype-ambiguity-main} visualizes this controlled ambiguity comparison.

\paragraph{Diagnostic selection.}
MovieLens tests the heterogeneous-item setting of Proposition~\ref{prop:covlatent} without assuming its latent model is exact. In ML1M $(2,80,.7)$, high-bin SeqTXCount has $.673$ NLL and $.696$ utility, versus $.629/.723$ for ANP and $.628/.719$ for FlowLatent. On the top-10\% slice, the baseline $.690/.689$ becomes $.609/.735$ for ANP and $.603/.729$ for FlowLatent. ML10M $(2,120,.6)$ gives $.670/.691$ for SeqTXCount versus $.639/.715$ and $.639/.712$, respectively. All comparisons use identical slates within each seed. The accompanying gain panels examine how the benefit changes across score bins under the fixed task, action cost, and shared evaluation protocol.

\begin{paperblock}
\begin{minipage}{0.985\linewidth}
\captionof{table}{Headline tail-decision results, averaged over five seeds. Lower NLL and higher utility are better. Every reported gain compares the indicated heads on an identical pre-specified evaluation slice.}
\label{tab:main-results}
\vspace{-0.35em}
\scriptsize
\renewcommand{\arraystretch}{0.94}
\setlength{\tabcolsep}{2.2pt}
\makebox[\linewidth][c]{\resizebox{0.88\linewidth}{!}{%
\begin{tabular}{@{}lllllll@{}}
\toprule
Setting & Slice & Baseline & NLL/Util. & Joint head & NLL/Util. & Gain \\
\midrule
Covtype $n=96$ & all & SeqTXCount & 0.537/-- & ObjSetTF & \strongnum{0.523/--} & \strongnum{+.014 NLL} \\
ML1M $(2,80,.7)$ & high TaxScore & SeqTXCount & 0.673/0.696 & ANP & \strongnum{0.629/0.723} & \strongnum{+.044/+.027} \\
ML1M $(2,80,.7)$ & top 10\% & SeqTXCount & 0.690/0.689 & ANP & \strongnum{0.609/0.735} & \strongnum{+.081/+.046} \\
ML10M $(2,120,.6)$ & high TaxScore & SeqTXCount & 0.670/0.691 & ANP & \strongnum{0.639/0.715} & \strongnum{+.031/+.024} \\
Budgeted switch & full test & SeqTXCount & -- & top 20--30\% & -- & \strongnum{+.006--.012 util.} \\
\bottomrule
\end{tabular}
}}
\end{minipage}
\end{paperblock}

\begin{paperblock}
\includegraphics[width=0.96\linewidth]{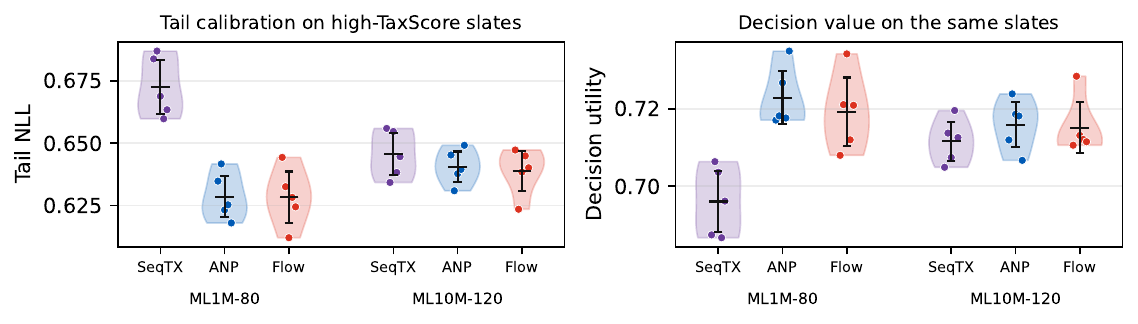}
\captionof{figure}{MovieLens decision-critical slate audit. Each point is one seed on the pre-specified high-TaxScore slice. Violins show the seed-level distribution and intervals show the mean with a 95\% standard-error band. Stochastic global-latent heads improve tail NLL and utility over direct Transformer count heads across both MovieLens scales and critical slices.}
\label{fig:main-movielens}
\end{paperblock}

Routing 20--30\% of slates yields $.021$--$.035$ utility gains on routed slates and $.006$--$.012$ overall: 2.1--3.5 versus 0.6--1.2 percentage points of normalized predict/defer payoff.

\begin{paperblock}
\includegraphics[width=0.77\linewidth]{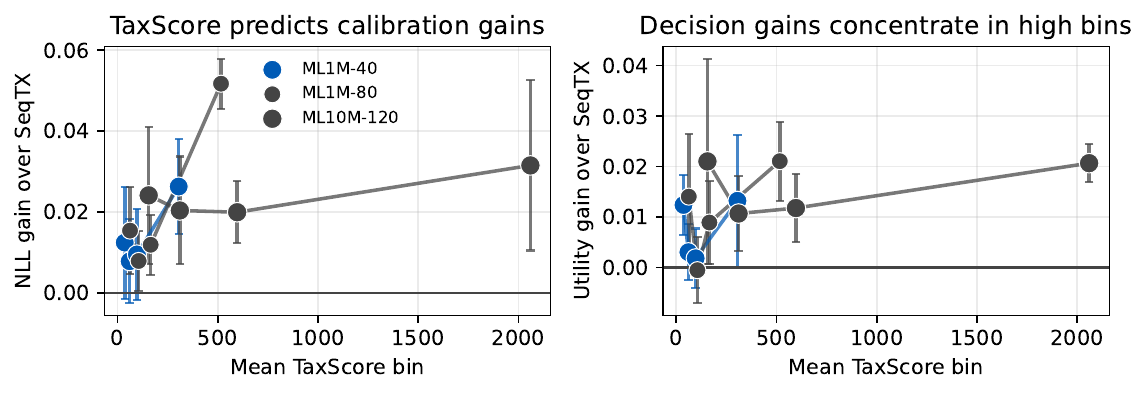}
\captionof{figure}{TaxScore gain diagnostics. The two panels compare realized NLL and utility gains across score bins, testing whether routing concentrates valuable joint inference.}
\label{fig:main-diagnostics}
\end{paperblock}
\vspace{-0.15em}

Appendix Table~\ref{tab:movie-proxy-robustness} perturbs candidate weights, margins, and budgets, retaining top-20\% NLL gains of $.027$--$.054$ on ML1M and $.039$--$.064$ on the bounded-user ML10M subset across six near-default variants. Switch-utility gains range from $.002$ to $.009$ and $-.004$ to $.003$, respectively. Single-signal selectors are weaker or unstable. Threshold perturbations retain positive ML1M high-slice NLL gains (Table~\ref{tab:movie-threshold-robustness}). This evidence supports local enrichment while showing how decision utility also depends on crossing the cost-specific action boundary on the selected evaluation slates.

\paragraph{Alternative explanations.}
On MovieLens, we hold the Transformer backbone, split, seeds, and decision rule fixed and evaluate SeqTXGlobal with $64$ components, matching the latent-evaluation budget of ANP/Flow. Relative to SeqTXCount, top-20\% NLL improves $.0228\pm.0118$ and utility $.0173\pm.0134$, both in five of five seeds. Full-population NLL increases by $.0032\pm.0057$. This paired comparison localizes the improvement to flagged groups and distinguishes the gain in their tail probabilities from the weaker population-wide result under the same matched-backbone protocol (Table~\ref{tab:matched-attribution}).

\paragraph{Averaging location.}
Holding the same three trained two-tower members fixed isolates where averaging occurs: EnsMean aggregates averaged item probabilities, whereas EnsMix averages the three tail probabilities. Both use $72{,}579$ parameters and the same member predictions. EnsMix reduces full-population NLL by $.0108\pm.0034$, Brier score by $.00066\pm.00026$, and ECE by $.00186\pm.00105$, with each probability metric improving in all five seeds. Utility changes by $-.0011\pm.0030$ overall and $-.0036\pm.0054$ on the top 20\%. Preserving the mixture improves probability quality in this control. The larger ANP/Flow utility gains also depend on the shared uncertainty represented by the corresponding learned posterior families (Table~\ref{tab:ensemble-attribution}).

\begin{paperblock}
\begin{minipage}{0.50\linewidth}
\centering
\includegraphics[width=0.91\linewidth]{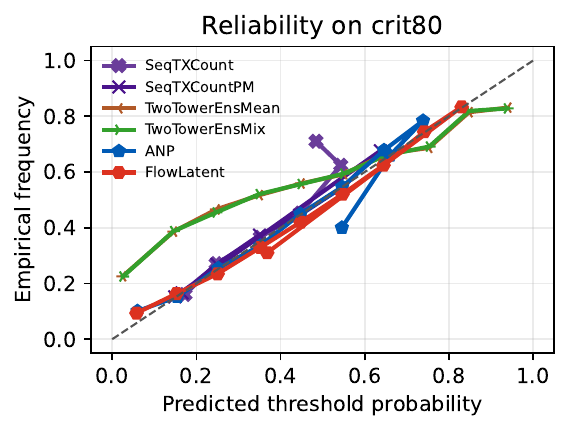}
\end{minipage}\hfill
\begin{minipage}{0.45\linewidth}
\centering
\includegraphics[width=0.91\linewidth]{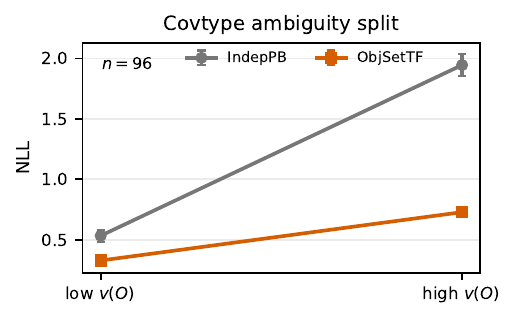}
\end{minipage}
\captionof{figure}{Calibration and mechanism checks. ML1M-80 reliability tests tail calibration. Covtype high-$v(O)$ prefixes expose the theorem-aligned ambiguity mechanism.}
\label{fig:covtype-ambiguity-main}
\end{paperblock}
\vspace{-0.2em}

\paragraph{Equal-budget routing.}
On ML1M-80, all selectors route exactly $330/1652$ or $496/1652$ test slates (20\% or 30\%). Baselines use SeqTXCount uncertainty $p(1-p)$ or average $100$ random masks per seed. Each mixed policy reselects its confidence threshold on validation users. Table~\ref{tab:routing-main} shows that TaxScore improves probability quality and payoff at each budget. Its paired advantage over each comparator is positive in all five seeds for both heads and metrics. This exact-budget study and the original validation-cutoff switch in Table~\ref{tab:main-results} use distinct operating points, so the interpretation of each estimate follows its declared selector, realized routing fraction, and policy-tuning rule.

\begin{paperblock}
\begin{minipage}{0.985\linewidth}
\captionof{table}{Equal-budget routing on ML1M-80. Each cell reports full-population NLL reduction / utility gain over SeqTXCount (five-seed means). All selectors route the same number of test slates. Appendix Table~\ref{tab:matched-routing} reports paired five-seed variability and the policy-tuning protocol.}
\label{tab:routing-main}
\centering\footnotesize
\setlength{\tabcolsep}{4pt}
\renewcommand{\arraystretch}{1.08}
\begin{tabular}{@{}lccc@{}}
\toprule
Rich head / fraction & TaxScore & Cheap uncertainty & Random \\
\midrule
ANP / 20\% & 0.0105 / 0.0071 & 0.0058 / 0.0024 & 0.0045 / 0.0032 \\
ANP / 30\% & 0.0137 / 0.0089 & 0.0085 / 0.0035 & 0.0066 / 0.0044 \\
FlowLatent / 20\% & 0.0106 / 0.0094 & 0.0050 / 0.0019 & 0.0037 / 0.0030 \\
FlowLatent / 30\% & 0.0127 / 0.0098 & 0.0068 / 0.0035 & 0.0058 / 0.0042 \\
\bottomrule
\end{tabular}
\end{minipage}
\end{paperblock}

\paragraph{Cost and latency.}
The archived five-seed cost sensitivity study retunes the confidence threshold at each $\lambda\in\{.10,.20,.30,.40,.50\}$. Its reported two-head-average gains over $.20$--$.50$ range from $.009$ to $.044$ on the high slice and $.002$ to $.011$ with 20\% routing. At $.10$, differences are within $3\times10^{-4}$ of zero as cheap deferral suppresses action changes (Appendix~\ref{app:cost-sensitivity}). Two NVIDIA L20 runs measure head-only online latency of $.58$--$.70$ ms for SeqTXCount, $5.92$--$8.07$ for ANP, and $6.48$--$8.76$ for Flow. Weighting these measurements by a 20--30\% routing fraction estimates $1.64$--$3.11$ ms/slate, 63--74\% below rich-head-only inference and $1.07$--$2.42$ ms above cheap-only inference. The head-only timing protocol excludes scoring, transfer, and queueing overhead (Table~\ref{tab:latency-audit}). The sensitivity archive retains summary ranges. Paired attribution and latency results are independently reaggregated from the released individual per-seed measurements and separately archived clean per-run timing records.

\paragraph{Boundaries and use.}
The controls support two levels of attribution. The fixed-member ensemble comparison isolates averaging location because member predictions are reused. The matched-backbone comparison also changes the learned output family, so optimization and approximation error remain part of its empirical effect. Together, they connect an exactly controlled operation to a trainable shared-latent alternative. Their agreement on flagged-slice probability gains strengthens the mechanism evidence while keeping the interpretation specific to the measured contrast.

The auxiliary tasks clarify where this mechanism matters. On Brown Creeper, essentially flat NLL and modest utility changes illustrate a weak-gain boundary. In contrast, the semi-synthetic Breast Cancer batch check shows event-TV increasing across oracle ambiguity bins (Table~\ref{tab:batch-inspection}), extending the controlled mechanism evidence beyond MovieLens. Together, these results favor selective inference when resolving shared uncertainty changes the downstream group-level decision.

\paragraph{Decision value and inference cost.}
The experiments reveal a useful distinction between probability quality and action value. NLL and Brier gains measure improvements in the tail posterior, supported by the reliability and ECE results. Utility gains arise when those improvements change a prediction or deferral at the chosen cost. This explains the fixed-ensemble comparison: preserving the mixture improves calibration, yet leaves the payoff nearly unchanged because many refined probabilities lead to the same action. Consequently, the strongest case for richer inference combines better probabilities on flagged groups with positive full-population switch gains at an affordable routing budget.

This distinction also guides adaptation to a new application. Candidate rules and score regularization can be fitted on development groups using the information available at decision time. With the utility specified, held-out validation groups then determine the routing fraction and confidence threshold that offer the best payoff within the compute budget. Freezing this policy before test evaluation yields a direct comparison with cheap-only inference. When the population or reward changes, renewed validation updates the routing rule to reflect where improved joint prediction can change the action.

The computational trade-off follows from the same policy. Suppose scoring costs $L_S$ and routes a fraction $r$ of groups to a rich head, with head latencies $L_{\rm cheap}$ and $L_{\rm rich}$. Executing only the selected head gives an expected latency of $L_S+(1-r)L_{\rm cheap}+rL_{\rm rich}$. Thus routing is faster than rich-head-only inference when
\[
L_S<(1-r)(L_{\rm rich}-L_{\rm cheap}).
\]
The measured head latencies quantify the right-hand side, leaving scoring, batching, and queueing overhead to the serving implementation. If the cheap head also runs on rich-routed groups, its extra pass must enter this accounting. Jointly measuring the resulting latency and payoff therefore identifies the routing budget that best balances the application's decision value and computational requirements.

\section{Conclusion}
Independent marginalization can preserve item means while changing the posterior of a hidden aggregate. This work unifies the approximation, its decision-specific consequences, and an empirical audit of when shared posterior computation is valuable. Classical stochastic orders support the count comparison, while threshold, covariance, and resolution analyses specify the information needed by the action. TaxScore guides selective inference under a declared utility and budget. Extending the audit to nonadditive combinatorial objectives, shifting populations, and end-to-end routing costs will test the value of selective joint prediction under broader objectives and operating conditions.

\section*{Acknowledgments}
This work was supported by the National Key R\&D Program of China (Grant No.\ 2025YFF0514701).

\clearpage
\bibliographystyle{plainnat}
\bibliography{refs}

\appendix

\clearpage
\noindent\textbf{Supplementary material.}
Appendix~\ref{app:deferred} gives formal consequences and notation, Appendix~\ref{app:proofs} contains proofs, and Appendix~\ref{app:expdetails} documents implementation and evaluation. Appendices~\ref{app:tables} and~\ref{app:figures} collect supplementary numerical results, empirical comparisons, and diagnostic figures.

\section{Formal Consequences and Robustness}
\label{app:deferred}

\begin{paperblock}
\begin{minipage}{0.985\linewidth}
\captionof{table}{Terminology and symbols. The main text uses standard probabilistic objects. The two named diagnostics below are defined through their mathematical role and available inputs.}
\label{tab:terms}
\small
\begin{tabular}{@{}p{.25\linewidth}p{.71\linewidth}@{}}
\toprule
Term or symbol & Meaning and role \\
\midrule
$O,W_{1:n},n$ & Observed prefix, visible hidden-item covariates, and hidden block size. All posterior comparisons condition on the available information. \\
$Q\mid O\sim\nu_O$ & Posterior of a shared Bernoulli success rate. Conditional on $Q$, hidden items are independent with probability $Q$. \\
$\bar q,v(O)$ & Posterior mean and variance of $Q$. $v(O)$ measures unresolved uncertainty in the shared success rate. \\
$R_n,\widehat R_n$ & Exact hidden count and count from mean-preserving independent completion. The latter uses a product of marginal predictions. \\
$G,\bar G_O,M_n$ & Directing random probability measure, its posterior mean, and the empirical measure of the hidden observations. \\
$\Theta,q_i(\Theta)$ & Shared latent state and heterogeneous conditional item success probabilities. \\
$Z,P,\widehat P$ & Decision outcome, target law $\Law(Z\mid O,W)$, and model predictive law conditional on the same information. \\
$a_P,a_{\widehat P}$ & Actions maximizing expected bounded utility under $P$ and $\widehat P$, respectively. \\
$\TV(P,\widehat P)$ & $\sup_A|P(A)-\widehat P(A)|$. Bounds each $[0,1]$ test-function error and gives an upper bound on Bayes regret. \\
$c,\rho,\Delta_{\rm count}$ & Tail threshold, per-item probability margin, and count margin. $\Delta_{\rm count}=n\rho$. \\
Serialization tax & Aggregate-posterior error induced by independent marginalization. Decision regret additionally depends on the utility. \\
Theoretical TaxScore & $n\rho_\pm^2$ in the two-point model, under Proposition~\ref{prop:taxscore}'s crossing conditions. \\
Empirical TaxScore $S(O)$ & Boundary-normalized grid-posterior ambiguity used to rank MovieLens slates. A surrogate whose routing value is evaluated empirically. \\
$\Psi$-probe compression & A summary determined only by finitely many posterior expectations of specified probes, including the constant function. \\
Posterior resolution & Ability of a grid, mixture, or samples to retain probability mass near the relevant boundary. Proposition~\ref{prop:resolution} gives a sufficient grid condition. \\
\bottomrule
\end{tabular}
\end{minipage}
\end{paperblock}

\paragraph{Formal implications and empirical hypotheses.}
Binary convex order directly compares expected convex count losses, including variance and stop-loss. The TV bound, equality conditions, threshold separation, and heterogeneous covariance identity have separate derivations. The empirical-measure result establishes a Laplace-functional comparison under its own exchangeability assumption. Finite-probe limits and resolution guarantees require their stated information and margin assumptions. The empirical audit evaluates learned-head calibration, utility, routing enrichment, and latency.

\paragraph{Additive and nonadditive actions.}
For any feasible action set independent of outcomes and utility $u(a,Y)=b(a)+\sum_i w_i(a)Y_i$, the conditional expected utility is $b(a)+\sum_iw_i(a)\E[Y_i\mid O,W]$. Any approximation preserving these marginal means leaves all action values, and hence the Bayes-optimal action set, unchanged. This includes fixed-position-weight ranking and linear top-$k$ selection, even with deterministic budget constraints. Joint-success probabilities, pairwise outcomes, joint coverage, and chance constraints generally require joint posterior information. They may be affected by dependence, but their analysis requires an explicit action and loss class with corresponding assumptions.

\begin{proposition}[Approximate de Finetti robustness]
\label{prop:perturb}
Let $n\ge2$. Let $R_n^\star$ be a Bernoulli-mixture surrogate count with latent posterior variance $v^\star(O)$, and suppose the actual remaining count $R_n$ satisfies
\[
\TV\!\left(\Law(R_n\mid O),\Law(R_n^\star\mid O)\right)\le \varepsilon_n(O).
\]
Let $\widehat R_n$ be any mean-preserving independent completion for the surrogate mean $\bar q^\star(O)$. Then
\[
\TV\!\left(\Law(R_n\mid O),\Law(\widehat R_n\mid O)\right)\ge v^\star(O)-\varepsilon_n(O).
\]
For every threshold $c\in(0,1)$, the threshold-posterior gap between $R_n$ and $\widehat R_n$ is at least the corresponding surrogate gap minus $\varepsilon_n(O)$.
\end{proposition}

\begin{corollary}[Finite-population robustness]
\label{cor:finitepop}
Suppose that, conditional on $O$, the target hidden block is a simple random sample without replacement of size $n$ from a finite exchangeable population of size $N_O\ge n$ with random success fraction $Q_N$. Let $R_n^\mathrm{wr}\mid Q_N,O\sim\mathrm{Bin}(n,Q_N)$ be the with-replacement surrogate. Then
\[
\TV\!\left(\Law(R_n\mid O),\Law(R_n^\mathrm{wr}\mid O)\right)
\le
\delta_{n,N_O}:=1-\frac{(N_O)_n}{N_O^n}
\le \frac{n(n-1)}{2N_O}.
\]
Consequently every TV lower bound for $n\ge2$, and every threshold lower bound, obtained for the surrogate degrades by at most $\delta_{n,N_O}$.
\end{corollary}

\begin{corollary}[Generating-function hierarchy]
\label{cor:pgf}
For every $z\ge 0$,
\begin{align}
\E[z^{R_n}\mid O]
&=
\E[(1-Q+Qz)^n\mid O] \label{eq:pgf-exact}\\
&\ge
\bigl(1-\bar q(O)+\bar q(O)z\bigr)^n \label{eq:pgf-iid}\\
&\ge
\prod_{i=t+1}^m \bigl(1-p_i(O)+p_i(O)z\bigr)
=
\E[z^{\widehat R_n}\mid O]. \label{eq:pgf-approx}
\end{align}
\end{corollary}

\begin{corollary}[Threshold-exceedance and factorial-moment hierarchy]
\label{cor:call}
\label{cor:factorial}
For every threshold $t\in\{0,\dots,n\}$,
\[
\E[(R_n-t)_+\mid O]\ge \E[(\widehat R_n-t)_+\mid O].
\]
For every $k\in\{2,\dots,n\}$,
\[
\E[(R_n)_k\mid O]
=
n_{(k)}\,\E[Q^k\mid O]
\ge
n_{(k)}\,\bar q(O)^k
\ge
\E[(\widehat R_n)_k\mid O].
\]
\end{corollary}

\begin{corollary}[TV gap and equality characterization]
\label{cor:tv}
\label{cor:equality}
Assume $n\ge2$. For every mean-preserving independent completion,
\begin{align}
\Var(R_n\mid O)
&=
n\,\bar q(O)\bigl(1-\bar q(O)\bigr)+n(n-1)v(O), \label{eq:var-exact}\\
\Var(\widehat R_n\mid O)
&\le
n\,\bar q(O)\bigl(1-\bar q(O)\bigr), \label{eq:var-approx}\\
\Var(R_n\mid O)-\Var(\widehat R_n\mid O)
&=
n(n-1)v(O)+\sum_{i=t+1}^m \bigl(p_i(O)-\bar q(O)\bigr)^2,
\label{eq:var-gap}
\end{align}
and
\[
\TV\!\left(\Law(R_n\mid O),\Law(\widehat R_n\mid O)\right)\ge v(O).
\]
Thus $\Law(R_n\mid O)=\Law(\widehat R_n\mid O)$ if and only if $v(O)=0$ and $p_i(O)=\bar q(O)$ for every hidden item. For $n=1$, both count laws are ${\rm Bernoulli}(\bar q(O))$, so the count-law discrepancy is identically zero.
\end{corollary}

\begin{proposition}[Binary stop/reveal rule]
\label{prop:reveal}
Fix a reveal cost $\lambda\in(0,\tfrac12)$ and actions $\{0,1,\mathrm{rev}\}$ with rewards
\[
u(a,y)=\mathbf{1}[a=y],
\qquad
u(\mathrm{rev},y)=1-\lambda,
\qquad
y\in\{0,1\}.
\]
If $p=\Prob(Y=1\mid O)$, the Bayes rule predicts $0$ for $p\le\lambda$, reveals for $\lambda<p<1-\lambda$, and predicts $1$ for $p\ge1-\lambda$. If $p$ lies at least $\eta>0$ inside one of these three decision regions while an approximate posterior selects a different region, the value loss is at least $\eta$.
\end{proposition}

\begin{corollary}[Stop-loss and CVaR-style tail risk]
\label{cor:stoploss}
Under Theorem~\ref{thm:universal}, for every threshold $t\in\{0,\dots,n\}$,
\[
\E[(R_n-t)_+\mid O]\ge \E[(\widehat R_n-t)_+\mid O],
\qquad
\E[(t-R_n)_+\mid O]\ge \E[(t-\widehat R_n)_+\mid O].
\]
Thus independent completion can understate stop-loss, CVaR-style, and chance-constrained tail risk whenever the scalar loss is an increasing affine function of the hidden count.
\end{corollary}

\begin{corollary}[Linear-statistic hierarchy]
\label{cor:linear}
Fix bounded measurable functions $g_1,\dots,g_d$ and define
\[
S_n=\bigl(M_n(g_1),\dots,M_n(g_d)\bigr),
\qquad
\widehat S_n=\bigl(\widehat M_n(g_1),\dots,\widehat M_n(g_d)\bigr).
\]
Then for every $\theta\in\mathbb{R}^d$,
\[
\E[e^{\theta^\top S_n}\mid O]
\ge
\left(\int \exp\!\left(\sum_{j=1}^d \theta_j g_j(x)\right)\,d\bar G_O(x)\right)^n
\ge
\E[e^{\theta^\top \widehat S_n}\mid O].
\]
\end{corollary}

\begin{corollary}[Finite-partition count-vector hierarchy]
\label{cor:partition}
Fix a measurable partition $\mathcal A=(A_1,\dots,A_d)$ and define
\[
P^\mathcal A=\bigl(G(A_1),\dots,G(A_d)\bigr),
\qquad
\bar p^\mathcal A(O)=\E[P^\mathcal A\mid O].
\]
Let $N_n^\mathcal A=(N_1,\dots,N_d)$ with $N_j=\sum_{i=t+1}^m \mathbf{1}[X_i\in A_j]$. Let $\widehat N_n^\mathcal A=\sum_{i=t+1}^m \widehat Y_i$ be any independent categorical completion with $\widehat Y_i\in\{e_1,\dots,e_d\}$ and $\sum_i \E[\widehat Y_i\mid O]=n\,\bar p^\mathcal A(O)$. Then for every $\theta\in\mathbb{R}^d$,
\[
\E[e^{\theta^\top N_n^\mathcal A}\mid O]
\ge
\left(\sum_{j=1}^d \bar p_j^\mathcal A(O)e^{\theta_j}\right)^n
\ge
\E[e^{\theta^\top \widehat N_n^\mathcal A}\mid O].
\]
Moreover, if $n\ge2$,
\begin{align*}
\Cov(N_n^\mathcal A\mid O)-\Cov(\widehat N_n^\mathcal A\mid O)
&\succeq n(n-1)\Cov(P^\mathcal A\mid O),\\
\TV\!\left(\Law(N_n^\mathcal A\mid O),\Law(\widehat N_n^\mathcal A\mid O)\right)
&\ge \tr\!\Cov(P^\mathcal A\mid O).
\end{align*}
\end{corollary}

\begin{corollary}[Finite-probe reductions]
\label{cor:permavg}
\label{cor:moment}
\label{cor:parametric}
\label{cor:countparam}
Finite permutation averages of rollout posteriors driven by the same probe vector, finite moment summaries, parametric latent posterior heads whose parameters are functions of finitely many probe expectations, and parametric count posterior heads whose parameters are functions of finitely many probe expectations are all $\Psi$-probe compressions. Hence Proposition~\ref{prop:probe} applies whenever its row-span condition holds. Concrete dictionaries include $\Psi(q)=(q,q^2,\dots,q^K)$ for moment heads, $\Psi(q)$ equal to the sufficient statistics used to fit a Beta or fixed exponential-family head, the finite rollout state statistics exposed to a count head, and the union of such statistics across a fixed number of permutation orders. The row-span test asks whether the vector $\mathbf{1}[q_j>c]$ on the witness support is linearly recoverable from that dictionary.
\end{corollary}

\section{Proofs}
\label{app:proofs}

\subsection{Proof of Theorem~\ref{thm:universal}}
\label{app:proof-universal}
\[
R_n\mid Q,O\sim \mathrm{Binomial}(n,Q).
\]
Fix a function $\phi:\{0,\dots,n\}\to\mathbb{R}$ and define
\[
B_\phi(q)=\E[\phi(\mathrm{Bin}(n,q))]
=
\sum_{r=0}^n \phi(r)\binom{n}{r}q^r(1-q)^{n-r}.
\]
The Bernstein second-derivative identity gives
\[
B_\phi''(q)
=
n(n-1)\sum_{r=0}^{n-2}\Delta^2\phi(r)\binom{n-2}{r}q^r(1-q)^{n-2-r},
\]
where $\Delta^2\phi(r)=\phi(r+2)-2\phi(r+1)+\phi(r)$. Hence if $\phi$ is discrete convex, then $B_\phi$ is convex on $[0,1]$.

Applying Jensen to the latent posterior gives the homogeneous exchangeable mixture comparison of \citet[Theorem~2.11]{cousin2008comparison}, here conditional on $O$:
\[
\E[\phi(R_n)\mid O]
=
\E[B_\phi(Q)\mid O]
\ge
B_\phi(\E[Q\mid O])
=
\E[\phi(\mathrm{Bin}(n,\bar q(O)))].
\]
On the other hand, \citet[Theorem~3]{hoeffding1956} showed that among all sums of independent Bernoulli random variables with a fixed mean, the mean-matched binomial is maximal in convex order. Therefore
\[
\E[\phi(\widehat R_n)\mid O]
\le
\E[\phi(\mathrm{Bin}(n,\bar q(O)))].
\]
Combining the two displays yields
\[
\E[\phi(R_n)\mid O]\ge \E[\phi(\widehat R_n)\mid O],
\]
which proves Theorem~\ref{thm:universal}.

\subsection{Proof of the one-dimensional corollaries}
For Corollary~\ref{cor:pgf}, choose $\phi(r)=z^r$ for any fixed $z\ge 0$. Then
\[
\Delta^2\phi(r)=z^r(z-1)^2\ge 0,
\]
so $\phi$ is discrete convex and Theorem~\ref{thm:universal} gives
\[
\E[z^{R_n}\mid O]\ge \E[z^{\widehat R_n}\mid O].
\]
The exact expression \eqref{eq:pgf-exact} follows from the conditional binomial law. The upper envelope \eqref{eq:pgf-iid} and the lower envelope \eqref{eq:pgf-approx} follow from the same Jensen arguments as above: $q\mapsto (1-q+qz)^n$ is convex, while $p\mapsto \log(1-p+pz)$ is concave. These are the two scalar functions used in the comparisons.

For Corollary~\ref{cor:call}, $\phi_t(r)=(r-t)_+$ has second difference $\mathbf1[r=t-1]\ge0$, hence is discrete convex.

Conditional on $Q$, the binomial factorial moments are
\[
\E[(R_n)_k\mid Q,O]=n_{(k)}Q^k.
\]
Averaging over $Q\mid O$ gives
\[
\E[(R_n)_k\mid O]=n_{(k)}\,\E[Q^k\mid O].
\]
Since $q\mapsto q^k$ is convex on $[0,1]$ for every $k\ge 2$, Jensen gives
\[
\E[Q^k\mid O]\ge \bar q(O)^k.
\]

For the independent completion,
\[
\E[(\widehat R_n)_k\mid O]=k!\,e_k\!\left(p_{t+1}(O),\dots,p_m(O)\right),
\]
where $e_k$ is the $k$th elementary symmetric polynomial. By the standard Maclaurin inequality for nonnegative vectors \citep{marshall2011majorization},
\[
e_k(p_1,\dots,p_n)\le \binom{n}{k}\left(\frac{1}{n}\sum_{i=1}^n p_i\right)^k
=
\binom{n}{k}\bar q(O)^k.
\]
Thus
\[
\E[(\widehat R_n)_k\mid O]\le n_{(k)}\,\bar q(O)^k,
\]
which proves the factorial-moment hierarchy.

For $k=2$,
\[
\E[(R_n)_2\mid O]
=
n(n-1)\E[Q^2\mid O]
=
n(n-1)\bigl(\bar q(O)^2+v(O)\bigr),
\]
while
\[
\E[(\widehat R_n)_2\mid O]
=
\sum_{i\neq j} p_i(O)p_j(O)
=
\left(\sum_i p_i(O)\right)^2-\sum_i p_i(O)^2
\le
n(n-1)\bar q(O)^2.
\]
Both laws have the same mean $n\bar q(O)$, so
\[
\Var(Y)=\E[(Y)_2]+\E[Y]-\E[Y]^2
\]
for any integer-valued random variable $Y$. Substituting $Y=R_n$ and $Y=\widehat R_n$ proves \eqref{eq:var-exact}, \eqref{eq:var-approx}, and \eqref{eq:var-gap}.

For Corollary~\ref{cor:equality}, note that
\[
\Var(\widehat R_n\mid O)
=
\sum_{i=t+1}^m p_i(O)\bigl(1-p_i(O)\bigr)
=
n\bar q(O)\bigl(1-\bar q(O)\bigr)-\sum_{i=t+1}^m \bigl(p_i(O)-\bar q(O)\bigr)^2.
\]
Combining this display with \eqref{eq:var-exact} yields
\[
\Var(R_n\mid O)-\Var(\widehat R_n\mid O)
=
n(n-1)v(O)+\sum_{i=t+1}^m \bigl(p_i(O)-\bar q(O)\bigr)^2.
\]
If both laws are equal, then their variances coincide, so both nonnegative terms on the right must vanish. Hence $v(O)=0$ and $p_i(O)=\bar q(O)$ for every hidden item. Conversely, these conditions imply $Q=\bar q(O)$ almost surely and $p_i(O)\equiv \bar q(O)$, so both laws are exactly $\mathrm{Binomial}(n,\bar q(O))$.

For $n=1$, every mean-preserving completion has the same Bernoulli count law as the exact predictive count, so no TV lower bound in terms of $v(O)$ can hold. For $n\ge2$, use the test function
\[
f(r)=\frac{r(r-1)}{n(n-1)}\in[0,1].
\]
Then
\[
\E[f(R_n)\mid O]-\E[f(\widehat R_n)\mid O]
=
\frac{\E[(R_n)_2\mid O]-\E[(\widehat R_n)_2\mid O]}{n(n-1)}
\ge
v(O).
\]
By the variational characterization
\[
\TV(P,Q)=\sup_{0\le g\le 1}\left|\E_P[g]-\E_Q[g]\right|,
\]
this implies
\[
\TV\!\left(\Law(R_n\mid O),\Law(\widehat R_n\mid O)\right)\ge v(O),
\]
which proves Corollary~\ref{cor:tv}.

\subsection{Proofs for the empirical-measure extension}
\label{app:proof-empirical}
Conditional on $G$,
\[
\E\!\left[e^{\int h\,dM_n}\middle|G,O\right]
=
\prod_{i=t+1}^m \int e^{h(x)}\,dG(x)
=
\left(\int e^{h(x)}\,dG(x)\right)^n.
\]
Taking conditional expectation over $G\mid O$ yields \eqref{eq:partition-exact}. Since $x\mapsto x^n$ is convex on $(0,\infty)$ and
\[
\E\!\left[\int e^{h(x)}\,dG(x)\middle|O\right]
=
\int e^{h(x)}\,d\bar G_O(x),
\]
Jensen gives \eqref{eq:partition-iid}.

For the independent completion,
\[
\E\!\left[e^{\int h\,d\widehat M_n}\middle|O\right]
=
\prod_{i=t+1}^m \int e^{h(x)}\,d\mu_i(x\mid O).
\]
Applying the arithmetic-geometric mean inequality,
\[
\prod_{i=t+1}^m \int e^{h(x)}\,d\mu_i(x\mid O)
\le
\left(
\frac{1}{n}\sum_{i=t+1}^m \int e^{h(x)}\,d\mu_i(x\mid O)
\right)^n
=
\left(\int e^{h(x)}\,d\bar G_O(x)\right)^n,
\]
which proves \eqref{eq:partition-approx}. The Laplace-functional version follows by taking $h=-f$.

Corollary~\ref{cor:linear} is immediate from Theorem~\ref{thm:partition} after taking
\[
h(x)=\sum_{j=1}^d \theta_j g_j(x).
\]

For Corollary~\ref{cor:partition}, apply Theorem~\ref{thm:partition} with
\[
h(x)=\sum_{j=1}^d \theta_j \mathbf{1}[x\in A_j].
\]
Then
\[
\int h\,dM_n=\theta^\top N_n^\mathcal A,
\qquad
\int h\,d\widehat M_n=\theta^\top \widehat N_n^\mathcal A,
\qquad
\int e^{h(x)}\,d\bar G_O(x)=\sum_{j=1}^d \bar p_j^\mathcal A(O)e^{\theta_j},
\]
which proves the count-vector mgf hierarchy.

For the covariance formula, conditional on $P^\mathcal A$,
\[
\Cov(N_n^\mathcal A\mid P^\mathcal A,O)
=
n\left(\mathrm{Diag}(P^\mathcal A)-P^\mathcal A(P^\mathcal A)^\top\right).
\]
Taking expectations and adding the covariance of the conditional mean gives
\[
\Cov(N_n^\mathcal A\mid O)
=
n\left(\mathrm{Diag}(\bar p^\mathcal A)-\bar p^\mathcal A(\bar p^\mathcal A)^\top\right)
+n(n-1)\Cov(P^\mathcal A\mid O).
\]
For the completion,
\[
\Cov(\widehat N_n^\mathcal A\mid O)
=
\sum_{i=t+1}^m \left(\mathrm{Diag}(p_i^\mathcal A)-p_i^\mathcal A(p_i^\mathcal A)^\top\right)
\]
and therefore
\[
\Cov(\widehat N_n^\mathcal A\mid O)
=
n\left(\mathrm{Diag}(\bar p^\mathcal A)-\bar p^\mathcal A(\bar p^\mathcal A)^\top\right)
-\sum_{i=t+1}^m \bigl(p_i^\mathcal A-\bar p^\mathcal A\bigr)\bigl(p_i^\mathcal A-\bar p^\mathcal A\bigr)^\top
\]
which is dominated by the mean-matched multinomial covariance. Subtracting yields the PSD lower bound.

For $n\ge2$, use the bounded test function
\[
f(x)=\frac{1}{n(n-1)}\sum_{j=1}^d x_j(x_j-1)\in[0,1].
\]
Under the exact law,
\[
\E[f(N_n^\mathcal A)\mid O]
=
\sum_{j=1}^d \E[(P_j^\mathcal A)^2\mid O].
\]
Under the completion,
\begin{align*}
\E[f(\widehat N_n^\mathcal A)\mid O]
&=
\frac{1}{n(n-1)}\sum_{j=1}^d \sum_{i\neq \ell} p_{ij}^\mathcal A(O)p_{\ell j}^\mathcal A(O)\\
&=
\frac{1}{n(n-1)}\sum_{j=1}^d \left(n^2(\bar p_j^\mathcal A(O))^2-\sum_{i=t+1}^m (p_{ij}^\mathcal A(O))^2\right)\\
&\le
\sum_{j=1}^d (\bar p_j^\mathcal A(O))^2,
\end{align*}
because $\sum_i (p_{ij}^\mathcal A)^2\ge n(\bar p_j^\mathcal A)^2$ for each coordinate. Hence
\[
\E[f(N_n^\mathcal A)\mid O]-\E[f(\widehat N_n^\mathcal A)\mid O]
\ge
\sum_{j=1}^d \Var(P_j^\mathcal A\mid O)
=
\tr\!\Cov(P^\mathcal A\mid O),
\]
and the variational characterization of TV proves the final claim of Corollary~\ref{cor:partition}.

\begin{proposition}[No unrestricted multivariate convex-order extension]
\label{prop:nocx}
There exist $d=3$, $n=2$, a convex function $\psi:\mathbb{R}^3\to\mathbb{R}$, and a posterior on $P\in\Delta_3$ with mean $\bar p$ such that, if
\[
N\mid P\sim \mathrm{Multinomial}(2,P),
\qquad
\bar N\sim \mathrm{Multinomial}(2,\bar p),
\]
then
\[
\E[\psi(N)]<\E[\psi(\bar N)].
\]
This counterexample rules out extending Theorem~\ref{thm:universal} to all convex functions of non-binary count vectors.
\end{proposition}

\begin{proof}
Work on the simplex
\[
T=\{(x,y)\in\mathbb{R}^2:\ x\ge 0,\ y\ge 0,\ x+y\le 1\},
\]
and define
\[
\phi(x,y)=\max\!\left\{-x-2y-\frac32,\ -x+\frac12 y-1,\ \frac32 x-y-2\right\}.
\]
This $\phi$ is convex as the maximum of affine functions. Set
\[
\psi(n_1,n_2,n_3)=\phi(n_1/2,n_2/2),
\]
which is convex on $\mathbb{R}^3$ because it is the composition of $\phi$ with a linear map. On the degree-$2$ Bernstein lattice,
\[
\phi(1,0)=\phi(0,1)=-\frac12,\qquad \phi(0,0)=-1,
\]
\[
\phi\!\left(\frac12,\frac12\right)=\phi\!\left(\frac12,0\right)=-\frac54,\qquad
\phi\!\left(0,\frac12\right)=-\frac34.
\]
Hence
\[
B_2\phi(p_1,p_2,p_3)
=
-\frac12 p_1^2-\frac12 p_2^2-p_3^2-\frac52 p_1p_2-\frac52 p_1p_3-\frac32 p_2p_3.
\]
Now let $P$ take the two values
\[
p^{(1)}=\left(\frac1{10},\frac45,\frac1{10}\right),
\qquad
p^{(2)}=(0,0,1)
\]
with probability $1/2$ each, so that
\[
\bar p=\E[P]=\left(\frac1{20},\frac25,\frac{11}{20}\right).
\]
If $N\mid P\sim\mathrm{Multinomial}(2,P)$ and $\bar N\sim\mathrm{Multinomial}(2,\bar p)$, then
\[
\E[\psi(N)\mid P=p]=B_2\phi(p),
\qquad
\E[\psi(\bar N)]=B_2\phi(\bar p).
\]
Direct substitution gives
\[
\E[\psi(N)]
=
\frac12 B_2\phi\!\left(\frac1{10},\frac45,\frac1{10}\right)
+\frac12 B_2\phi(0,0,1)
=
-\frac{21}{25},
\]
whereas
\[
\E[\psi(\bar N)]
=
B_2\phi\!\left(\frac1{20},\frac25,\frac{11}{20}\right)
=
-\frac{333}{400}.
\]
Since $-\frac{21}{25}<-\frac{333}{400}$, the claimed failure follows.
\end{proof}

For Proposition~\ref{prop:perturb}, the triangle inequality gives
\[
\TV\!\left(\Law(R_n\mid O),\Law(\widehat R_n\mid O)\right)
\ge
\TV\!\left(\Law(R_n^\star\mid O),\Law(\widehat R_n\mid O)\right)-\varepsilon_n(O).
\]
Applying Corollary~\ref{cor:tv} to the surrogate law yields the first inequality. For the threshold event, the difference between the actual and surrogate event probabilities is bounded by
\[
\left|
\Prob(R_n\ge \lceil cn\rceil\mid O)-\Prob(R_n^\star\ge \lceil cn\rceil\mid O)
\right|
\le
\TV\!\left(\Law(R_n\mid O),\Law(R_n^\star\mid O)\right)
\le
\varepsilon_n(O),
\]
and rearranging proves the second display.

\subsection{Proof of Proposition~\ref{prop:covlatent}}
\label{app:proof-covlatent}
All probabilities below are conditional on $(O,W_{1:n})$. Conditional independence given $\Theta$ gives
\[
\E[R(R-1)]
=
\sum_{i\ne j}\E[q_i(\Theta)q_j(\Theta)],
\qquad
\E[\widehat R(\widehat R-1)]
=
\sum_{i\ne j}\bar q_i\bar q_j,
\]
which proves the factorial-moment identity. Also
\[
\Var(R)
=
\sum_i \bar q_i(1-\bar q_i)
+
\sum_{i\ne j}\Cov(q_i(\Theta),q_j(\Theta)),
\]
because $\E[q_i(1-q_i)]+\Var(q_i)=\bar q_i(1-\bar q_i)$ for each coordinate. The independent completion has variance $\sum_i\bar q_i(1-\bar q_i)$, proving the variance identity.

Association gives nonnegative covariance for every pair $q_i(\Theta),q_j(\Theta)$ because each is coordinatewise nondecreasing. For the tail bound, condition on $\Theta\in\mathcal H$. The heterogeneous Bernoulli sum has conditional mean at least $a+\Delta_{\rm count,\mathcal H}$. Hoeffding's inequality for independent bounded variables gives, for every such $\Theta$,
\[
\Prob(R<a\mid \Theta,O,W)\le e^{-2\Delta_{\rm count,\mathcal H}^2/n}.
\]
Therefore $\Prob(R\ge a\mid O,W)\ge \alpha(1-e^{-2\Delta_{\rm count,\mathcal H}^2/n})$. The independent completion has mean at most $a-\Delta_{\rm count,ind}$, so Hoeffding gives
\[
\Prob(\widehat R\ge a\mid O,W)\le e^{-2\Delta_{\rm count,ind}^2/n}.
\]
Subtracting the two displays proves the claim.

\subsection{Proof of Proposition~\ref{prop:signedcov}}
The first display is the covariance identity in Proposition~\ref{prop:covlatent} after writing each off-diagonal covariance as $C_{ij}=C_{ij,+}-C_{ij,-}$ and summing over $i\ne j$. The lower bound follows immediately from $S_-\le \eta S_+$. The posterior-sample audit is the empirical covariance matrix of the square-integrable random vector $(q_1(\Theta),\dots,q_n(\Theta))$, establishing consistency by the law of large numbers.

\subsection{Proof of Corollary~\ref{cor:finitepop}}
Condition on the finite population and draw $n$ indices with replacement from $\{1,\dots,N_O\}$. If no index repeats, the ordered sample has the same law as sampling without replacement. Hence the total variation distance between the with- and without-replacement ordered samples is at most the collision probability
\[
\delta_{n,N_O}=1-\frac{(N_O)_n}{N_O^n}.
\]
The count is a measurable function of the ordered sample, so the same TV bound holds for the count laws. Finally,
\[
1-\prod_{j=0}^{n-1}\left(1-\frac{j}{N_O}\right)
\le
\sum_{j=0}^{n-1}\frac{j}{N_O}
=\frac{n(n-1)}{2N_O},
\]
and Proposition~\ref{prop:perturb} transfers the surrogate lower bounds with this value of $\varepsilon_n(O)$.

\subsection{Proof of Theorem~\ref{thm:decision}}
All expectations in this proof are conditional on the same observed information $(O,W)$. For any fixed action $a$,
\[
\left|\E_P[u(a,Z)]-\E_{\widehat P}[u(a,Z)]\right|\le \TV(P,\widehat P),
\]
since $u(a,\cdot)\in[0,1]$ and TV controls the expectation of every bounded test function. Therefore
\begin{align*}
\E_P[u(a_P,Z)]-\E_P[u(a_{\widehat P},Z)]
&=
\Big(\E_P[u(a_P,Z)]-\E_{\widehat P}[u(a_P,Z)]\Big)\\
&\quad+
\Big(\E_{\widehat P}[u(a_P,Z)]-\E_{\widehat P}[u(a_{\widehat P},Z)]\Big)\\
&\quad+
\Big(\E_{\widehat P}[u(a_{\widehat P},Z)]-\E_P[u(a_{\widehat P},Z)]\Big)\\
&\le \TV(P,\widehat P)+0+\TV(P,\widehat P),
\end{align*}
because $a_{\widehat P}$ is optimal under $\widehat P$. Nonnegativity is immediate from optimality of $a_P$ under $P$.

\paragraph{Scope and non-Bayes decisions.}\label{app:decision-scope}
The bound measures conditional regret under the target posterior. It averages over hidden outcomes and supplies an upper bound for the specified utility. For example, with binary classification utility, $P(Z=1)=.9$ and $\widehat P(Z=1)=.6$ have TV distance $.3$ but the same optimal action and zero regret. For an implemented action $\widetilde a$, define its departure from Bayes optimality under $\widehat P$ as
\[
\epsilon_{\widehat P}(\widetilde a)=\sup_a\E_{\widehat P}u(a,Z)-\E_{\widehat P}u(\widetilde a,Z)\ge0.
\]
The same proof gives regret at most $2\TV(P,\widehat P)+\epsilon_{\widehat P}(\widetilde a)$. The validation-selected confidence threshold used in MovieLens can differ from the analytic Bayes threshold. The experiment measures realized utility alongside predictive error. For a binary tail indicator $Z$, TV is the absolute tail-probability error. Projecting a count law to the binary event contracts TV, preserving the count-level upper bound.

\subsection{Proof of Proposition~\ref{prop:reveal}}
Under posterior $p=\Prob(Y=1\mid O)$, the three Bayes values are
\[
V_0(p)=1-p,
\qquad
V_1(p)=p,
\qquad
V_{\mathrm{rev}}(p)=1-\lambda.
\]
Hence action $0$ is optimal exactly when $1-p\ge 1-\lambda$ and $1-p\ge p$, which reduces to $p\le \lambda$ because $\lambda<\tfrac12$. By symmetry, action $1$ is optimal exactly when $p\ge 1-\lambda$, and otherwise reveal is optimal.

For the regret bounds, write
\[
V^\star(p)=\max\{1-p,p,1-\lambda\}.
\]
If $p\in[\lambda+\eta,1-\lambda-\eta]$ and $\widehat p\notin(\lambda,1-\lambda)$, then the approximate posterior chooses $0$ or $1$, whose value is at most $\max\{1-p,p\}\le 1-\lambda-\eta$, while $V^\star(p)=1-\lambda$. Hence the regret is at least $\eta$.

If $p\le \lambda-\eta$ and $\widehat p>\lambda$, the approximate posterior chooses action $1$ or reveal. Any other action has value at most $1-\lambda$, while $V^\star(p)=1-p\ge 1-\lambda+\eta$, so the regret is at least $\eta$. The symmetric argument handles $p\ge 1-\lambda+\eta$ and $\widehat p<1-\lambda$ with the same regret bound.

\subsection{Proof of Corollary~\ref{cor:stoploss}}
The functions $x\mapsto (x-t)_+$ and $x\mapsto (t-x)_+$ are convex on $\{0,\dots,n\}$, so Theorem~\ref{thm:universal} gives both stop-loss inequalities. For a scalar loss $L$, the standard representation
\[
\mathrm{CVaR}_\beta(L)=\inf_{s\in\mathbb R}\left\{s+\frac{1}{1-\beta}\E[(L-s)_+]\right\}
\]
shows that ordering every upper stop-loss term orders the CVaR objective for affine increasing functions of the hidden count. The buffer claim follows by applying this inequality to threshold slack.

\subsection{Worst-case tightness}
The factor $2$ in Theorem~\ref{thm:decision} is tight under the stated $[0,1]$ utility domain. Fix any $\delta\in(0,1/2]$ and $\varepsilon\in(0,\delta)$. Let $Z\in\{0,1\}$, let
\[
\widehat P(Z=1)=\frac12-\varepsilon,
\qquad
P(Z=1)=\frac12-\varepsilon+\delta,
\]
so $\TV(P,\widehat P)=\delta$. Consider binary classification utility $u(a,z)=\mathbf{1}[a=z]$. Under $\widehat P$, action $0$ is uniquely optimal. Under $P$, action $1$ is optimal whenever $\delta>\varepsilon$. If the approximate posterior selects the $\widehat P$-Bayes action and is evaluated under $P$, the regret is
\[
P(Z=1)-P(Z=0)=2\delta-2\varepsilon.
\]
Letting $\varepsilon\downarrow0$ proves sharpness of $2\TV(P,\widehat P)$ within the stated bounded-utility class.

\subsection{Proof of the asymptotic two-point separation}
\label{app:proof-taxscore}
Under the exact posterior,
\[
\Prob(Y_n(c)=1\mid O)
=
\alpha\,\Prob(\mathrm{Bin}(n,q_1)\ge \lceil c n\rceil)
+
(1-\alpha)\Prob(\mathrm{Bin}(n,q_0)\ge \lceil c n\rceil).
\]
Since $q_0<c<q_1$, the weak law of large numbers implies
\[
\Prob(\mathrm{Bin}(n,q_1)\ge \lceil c n\rceil)\to 1,
\qquad
\Prob(\mathrm{Bin}(n,q_0)\ge \lceil c n\rceil)\to 0.
\]
Hence the exact posterior converges to $\alpha$. Local completion replaces the mixture by $\mathrm{Bin}(n,\bar q)$, which converges to $\mathbf{1}[\bar q>c]$ for $\bar q\neq c$ by applying the weak law to this single mean-matched binomial.

\subsection{Proof of Corollary~\ref{cor:thresholdrate}}
If $\bar q<c$, then Hoeffding's inequality gives
\[
\Prob(\mathrm{Bin}(n,q_1)<\lceil cn\rceil)\le e^{-2n(q_1-c)^2},
\qquad
\Prob(\mathrm{Bin}(n,\bar q)\ge \lceil cn\rceil)\le e^{-2n(c-\bar q)^2}.
\]
Therefore
\[
p_n
=
\alpha\,\Prob(\mathrm{Bin}(n,q_1)\ge \lceil cn\rceil)
+
(1-\alpha)\Prob(\mathrm{Bin}(n,q_0)\ge \lceil cn\rceil)
\ge
\alpha\bigl(1-e^{-2n(q_1-c)^2}\bigr),
\]
while $\widehat p_n\le e^{-2n(c-\bar q)^2}$. Subtracting yields the first inequality.

If $\bar q>c$, then Hoeffding's inequality gives
\[
\Prob(\mathrm{Bin}(n,\bar q)<\lceil cn\rceil)\le e^{-2n(\bar q-c)^2},
\qquad
\Prob(\mathrm{Bin}(n,q_0)\ge \lceil cn\rceil)\le e^{-2n(c-q_0)^2}.
\]
Hence
\[
\widehat p_n\ge 1-e^{-2n(\bar q-c)^2},
\qquad
p_n\le \alpha+(1-\alpha)e^{-2n(c-q_0)^2},
\]
Subtracting these two inequalities gives the claimed upper-side threshold separation.

\subsection{Proof of Proposition~\ref{prop:taxscore}}
If $\bar q<c$, then $\rho_-(O;c)\le q_1-c$ and $\rho_-(O;c)\le c-\bar q$. Corollary~\ref{cor:thresholdrate} gives
\[
p_n-\widehat p_n
\ge
\alpha-\alpha e^{-2n(q_1-c)^2}-e^{-2n(c-\bar q)^2}
\ge
\alpha-(1+\alpha)e^{-2n\rho_-^2(O;c)}.
\]
The case $\bar q>c$ is the same:
\[
\widehat p_n-p_n
\ge
(1-\alpha)-e^{-2n(\bar q-c)^2}-(1-\alpha)e^{-2n(c-q_0)^2}
\ge
(1-\alpha)-2e^{-2n\rho_+^2(O;c)}.
\]
If $\rho_\pm(O;c)=\kappa\sqrt{v(O)}$, then $\mathsf{TaxScore}_n=n\kappa^2v(O)$, giving the claimed $n v(O)$ scale.

\subsection{Proof of Proposition~\ref{prop:resolution}}
Write $B_\gamma=\{|Q-c|\le\gamma\}$. The contribution of $B_\gamma$ to $|p_n^\star-p_{n,h}|$ is at most $\nu(B_\gamma)$ because both tail probabilities lie in $[0,1]$. On $Q\ge c+\gamma$, the projection satisfies $\pi_h(Q)\ge c+\gamma-h>c$. Hoeffding's inequality gives
\[
1-T_n(Q)\le e^{-2n\gamma^2},
\qquad
1-T_n(\pi_h(Q))\le e^{-2n(\gamma-h)^2},
\]
so $|T_n(Q)-T_n(\pi_h(Q))|\le 2e^{-2n(\gamma-h)^2}$. The case $Q\le c-\gamma$ is symmetric. Integrating over $B_\gamma^c$ proves the bound.

The final statement follows from the same decomposition: if $\nu(B_\gamma)$ is uncontrolled, the right-hand side may be order one for every $\gamma>h$, regardless of how small $h$ is relative to the rest of the state space. A boundary-mass condition is therefore needed for a useful rate.

\subsection{Proof of Proposition~\ref{prop:probe}}
\label{app:proof-probe}
Let the witness support be $\mathcal X=\{x_1,\dots,x_J\}\subset[0,1]\setminus\{c\}$, let $M_{\ell j}=\psi_\ell(x_j)$ include the constant-probe row, and let $b_j=\mathbf1[x_j>c]$. By assumption, $b\in\mathbb{R}^{J}$ lies outside the row span of $M$. Therefore there exists a vector $a\in\ker(M)$ such that
\[
b^\top a\neq 0.
\]

Write $a=a^+-a^-$ with disjoint nonnegative parts. Because the constant-probe row of $M$ is all ones and $Ma=0$,
\[
\sum_{j=1}^{J} a_j=0,
\]
so the positive and negative masses agree:
\[
A:=\sum_{j=1}^{J} a_j^+=\sum_{j=1}^{J} a_j^- > 0.
\]
Define two probability measures supported on $\mathcal X$ by
\[
\nu=\sum_{j=1}^{J}\frac{a_j^+}{A}\delta_{x_j},
\qquad
\nu'=\sum_{j=1}^{J}\frac{a_j^-}{A}\delta_{x_j}.
\]
For every probe $\psi_\ell$,
\[
\int \psi_\ell(q)\,d\nu(q)-\int \psi_\ell(q)\,d\nu'(q)
=
\frac{1}{A}\sum_{j=1}^{J} a_j \psi_\ell(x_j)
=
0,
\]
because the corresponding probe row of $M$ annihilates $a$. Moreover,
\[
\nu((c,1])-\nu'((c,1])
=
\frac{1}{A}\sum_{j=1}^{J} a_j \mathbf{1}[x_j>c]
=
\frac{1}{A} b^\top a
\neq 0.
\]
Set $\delta_c=|\nu((c,1])-\nu'((c,1])|>0$ for later reference.

Because both supports avoid $c$, the weak law implies
\[
\Prob_\nu(Y_n(c)=1)\to \nu((c,1]),
\qquad
\Prob_{\nu'}(Y_n(c)=1)\to \nu'((c,1]).
\]
A $\Psi$-probe compression produces the same approximate posterior on both instances, since all probe expectations agree. Writing the common approximate threshold probability as $\widehat p_n$, we obtain
\[
\max\bigl\{|\widehat p_n-\Prob_\nu(Y_n(c)=1)|,\ |\widehat p_n-\Prob_{\nu'}(Y_n(c)=1)|\bigr\}
\ge \frac{1}{2}\left|\Prob_\nu(Y_n(c)=1)-\Prob_{\nu'}(Y_n(c)=1)\right|.
\]
Taking $\limsup$ proves the posterior-error claim.

For the final sentence, fix the worse of the two instances in the previous display. On that instance, the exact posterior over $Y_n(c)$ is a Bernoulli law and the approximate posterior is another Bernoulli law with parameter error at least
\[
\frac{1}{2}\delta_c-o(1).
\]
For Bernoulli laws, this parameter error is exactly the TV gap. Applying the worst-case tightness construction after Theorem~\ref{thm:decision} on the binary event space therefore yields, for every $\varepsilon>0$, a bounded-utility decision problem whose asymptotic Bayes value loss is at least
\[
\delta_c-\varepsilon.
\]

\subsection{Proof of Corollary~\ref{cor:minimax}}
Proposition~\ref{prop:probe} already constructs two posterior laws $\nu,\nu'\in\mathcal V(\mathcal X)$ with matched probe expectations and distinct threshold masses whenever the row-span condition holds, so $\Delta_{\Psi,\mathcal X}(c)>0$.

Fix any $\Psi$-probe compression. For any pair $\nu,\nu'\in\mathcal V(\mathcal X)$ with matched probe expectations, the compression produces the same approximate posterior on both instances. Writing the common asymptotic threshold estimate as $\widehat p_n$, we have directly
\[
\max\bigl\{|\widehat p_n-\Prob_\nu(Y_n(c)=1)|,\ |\widehat p_n-\Prob_{\nu'}(Y_n(c)=1)|\bigr\}
\ge
\frac12\left|\Prob_\nu(Y_n(c)=1)-\Prob_{\nu'}(Y_n(c)=1)\right|.
\]
Taking $\limsup$ and then the supremum over all admissible pairs gives
\[
\sup_{\nu\in\mathcal V(\mathcal X)}
\limsup_{n\to\infty}
\left|
\widehat\Prob_\nu(Y_n(c)=1)-\Prob_\nu(Y_n(c)=1)
\right|
\ge
\frac12\Delta_{\Psi,\mathcal X}(c).
\]
The decision-loss statement follows by applying the worst-case tightness construction after Theorem~\ref{thm:decision} to the harder instance of a near-maximizing indistinguishable pair of posterior measures.

\subsection{Proofs of finite-probe reductions}
\paragraph{Permutation averages.}
Each rollout posterior $\Law(\widehat R_n^{(\pi)}\mid O)$ is, by assumption, a measurable function of the same probe vector $(s_1(O),\dots,s_K(O))$. A finite average of such functions is again a measurable function of that vector. The averaged predictor is consequently covered by Proposition~\ref{prop:probe}.

\paragraph{Finite moments.}
Choose distinct points
\[
0\le x_1<c<x_2<\cdots<x_{K+2}\le 1.
\]
For the moment probes $\psi_j(q)=q^j$, the row span of $M$ is exactly the set of value vectors obtained by evaluating polynomials of degree at most $K$ on these $K+2$ points. The threshold vector is
\[
b=(0,1,\dots,1).
\]
If $b$ lay in the row span of $M$, there would exist a polynomial $p$ of degree at most $K$ such that
\[
p(x_1)=0,
\qquad
p(x_j)=1 \ \text{for } j=2,\dots,K+2.
\]
Then the polynomial $p-1$ would have $K+1$ distinct roots $x_2,\dots,x_{K+2}$, forcing $p\equiv 1$, which contradicts $p(x_1)=0$. Thus $b$ lies outside the row span, so Proposition~\ref{prop:probe} applies. Combining this observation with Corollary~\ref{cor:permavg} proves the permutation-average extension.

\paragraph{Low-parameter latent heads.}
For each $r\in\{0,\dots,n\}$ define
\[
g_r(\theta)=\int_0^1 \binom{n}{r}q^r(1-q)^{n-r}\,d\nu_\theta(q).
\]
By measurability of the family $\{\nu_\theta\}$, each $g_r$ is measurable on $\Theta$. Therefore
\[
\Law(\widehat R_n\mid O)(r)=g_r\!\left(H\bigl(s_1(O),\dots,s_K(O)\bigr)\right)
\]
is a measurable function of the same probe vector $(s_1(O),\dots,s_K(O))$. This expresses the entire count law through the fixed probes, precisely the information restriction of Proposition~\ref{prop:probe}.

\paragraph{Low-parameter count heads.}
For each $n$ and $r\in\{0,\dots,n\}$, the coordinate map
\[
\theta\mapsto \mu_{n,\theta}(r)
\]
is measurable by assumption. Therefore each coordinate of
\[
\Law(\widehat R_n\mid O)=\mu_{n,\theta_n(O)}
\]
is a measurable function of the same probe vector $(s_1(O),\dots,s_K(O))$, since $\theta_n(O)=H_n(s_1(O),\dots,s_K(O))$. The resulting count law is thus a $\Psi$-probe compression covered by Proposition~\ref{prop:probe}.

\section{Experimental Protocols and Implementation Details}
\label{app:expdetails}
\label{app:exp}
\paragraph{Compute and aggregation protocol.}
The main benchmark runs used a Linux environment with NVIDIA L20 GPUs and a fixed software environment across seeds. Per-seed JSON records contain configuration, metrics, and audit metadata. Deterministic scripts generate the tables and figures. The added attribution controls and two clean latency runs retain their own raw records and protocols. The cost-sensitivity evidence is an archived summary. Its reproducibility limits are stated separately below.

\paragraph{Reproducibility package.}
The supplementary material includes paper sources, experiment code, per-seed and aggregated outputs, plotting scripts, and reproduction instructions for the environment, seed list, benchmark commands, and aggregation commands. External datasets are obtained from scikit-learn \citep{pedregosa2011scikit}, MovieLens \citep{harper2015movielens}, UCI \citep{blackard1998covertype,wolberg1995breast,alpaydin1998digits}, and the public MIL sources cited in each benchmark description, with download locations and preprocessing steps recorded in the corresponding scripts.

\paragraph{Breast-cancer benchmark.}
We used the Wisconsin Breast Cancer dataset \citep{wolberg1995breast} as exposed by scikit-learn \citep{pedregosa2011scikit}, split the raw examples into train/validation/test partitions, and generated hierarchical bags with $(m,k,q_0,q_1)=(5,2,0.2,0.8)$. The reported results use $5000$ train bags, $1200$ validation bags, and $1200$ test bags per seed.

\paragraph{Digits benchmark.}
We binarized the optical handwritten digits dataset \citep{alpaydin1998digits} as exposed by scikit-learn \citep{pedregosa2011scikit} by whether the digit lies in $\{5,6,7,8,9\}$, then generated hierarchical bags with $(m,q_0,q_1)=(7,0.2,0.8)$. Stop/reveal thresholds were tuned on validation bags over the grid $\{0.55,0.60,\dots,0.90\}$. The point metric in Table~\ref{tab:digits} uses reveal cost $0.06$, while the underlying policy sweep evaluates reveal costs in $\{0.00,0.02,\dots,0.12\}$. Per seed, the split contains $4000$ training bags and $1000$ bags each for validation and final evaluation.

\paragraph{Covtype benchmark.}
We used the full Forest CoverType dataset \citep{blackard1998covertype} as exposed by scikit-learn \citep{pedregosa2011scikit}, treated cover type $2$ as the positive class for binarization, and generated hierarchical bags with $(m,k,q_0,q_1)=(21,5,0.2,0.8)$. To keep the item-level baseline computationally stable at this scale, the raw logistic item model was fit on a random $100{,}000$-example subsample of the training split, while the bag-level models were fit on $12{,}000$ train bags per seed. The reported results use $3000$ validation bags and $3000$ test bags per seed.

\paragraph{Covtype threshold benchmark.}
Using the same raw covtype split, we generated threshold-risk bags with observed prefix length $k=3$ and hidden sizes $n\in\{16,48,96\}$. The label is the tail event
\[
\mathbf{1}\!\left[\sum_{i=k+1}^{k+n} Z_i \ge \left\lceil \frac{n}{2}\right\rceil\right].
\]
Each seed uses $8000$ train bags, $2000$ validation bags, and $2000$ test bags for each hidden size. The same item-level logistic model is reused across the sweep, and the bag-level predictors are retrained separately for each hidden size. The ARRollout baseline is trained as a Mittal-style multi-step predictor \citep{NEURIPS2025_e31fc00e} over hidden-item reveals and then marginalized back to the threshold event. ObjSetTF uses the same object-space task but replaces the object MLP with a Set Transformer.

\paragraph{Covtype active-learning benchmark.}
For active learning, we fix the same threshold-risk task at hidden size $n=96$. Each queried bag reveals its full latent item labels, after which the corresponding predictor is retrained and used to score the remaining pool by threshold-event uncertainty. We report budgets $\{256,512,1024,2048\}$ over a pool of $6144$ bags, with $1000$ validation and $2000$ test bags per seed.

\paragraph{Musk stop/reveal benchmark.}
We used the Musk multiple-instance benchmark \citep{dietterich1997solving}, where each molecule is a bag of conformations with a natural bag label. We kept molecules with at least $6$ conformations, formed train/validation/test partitions by molecule identity, and sampled fixed-size sub-bag views of size $m=6$. Stop/reveal thresholds were tuned on validation views over the grid $\{0.55,0.60,\dots,0.90\}$, and Appendix Figure~\ref{fig:musk} sweeps reveal cost in $\{0.00,0.02,\dots,0.10\}$.

\paragraph{Birds stop/reveal benchmark.}
We used the Brown Creeper multiple-instance benchmark \citep{briggs2012rank}, where each bag is an audio recording represented by local segments and the bag label indicates species presence. We kept bags with at least $16$ segments, split by bag identity, and sampled fixed-size views of size $m=16$. This natural grouped benchmark tests whether posterior-sensitive stop/reveal decisions remain informative beyond the controlled two-point latent model. Threshold tuning uses the Musk validation grid. Figure~\ref{fig:birds} reports the reveal-cost sweep $\{0.00,0.02,\dots,0.10\}$.

For the diagnostic natural-task sweep we also used MovieLens 100K \citep{harper2015movielens}. Each user is a group, each rated movie is an item, and the sampled-view label is whether at least half of the unrevealed slate receives a high rating ($\ge4$ stars) after observing three rated items from the same user. Item features contain movie genres and the observed rating/positive bit for the revealed prefix. No synthetic user latent is sampled. This task is closest to recommender triage among the natural sweeps. Gains concentrate in the high-TaxScore slice of these natural grouped decisions.

\paragraph{MovieLens tail-risk benchmarks.}
\label{app:movie-score}
We used MovieLens 1M \citep{harper2015movielens} for the main recommender slate experiment and MovieLens 10M as a scale-up check. Users with at least $k+n$ ratings were split by user identity into train/validation/test sets. For each sampled view, $k$ ratings are revealed and $n$ held-out movies form the hidden slate. The label is whether at least $\lceil cn\rceil$ hidden movies receive rating at least four stars. Movie genres and training-set movie popularity are visible for hidden items. Hidden ratings are reserved for outcome evaluation. Additional random 1M configurations $(5,50,0.7)$, $(2,100,0.7)$, and $(2,100,0.8)$ were used for full-distribution stress checks outside the headline decision-critical audit. The 10M scale-up uses the pre-specified cold-slate setting $(2,120,0.6)$ and a pool of 5000 eligible users before user-level splitting.

The decision-critical configurations $(2,40,0.7)$, $(2,80,0.7)$, and $(2,80,0.75)$ use a pre-specified triage rule. For each candidate view, compute the hidden-slate prior success rate using only training-set movie popularity logits and compute the observed prefix success rate using only the $k$ revealed ratings. Accept immediately if the hidden prior is within $0.08$ of $c$ and the observed prefix contains both outcomes. Otherwise keep the best of $80$ candidates under
\[
\left|\bar p_{\mathrm{hidden,prior}}-c\right|+0.20\left|\bar y_{\mathrm{obs}}-\frac12\right|.
\]
The candidate rule never reads hidden ratings, hidden positives, future outcomes, or predictions from evaluated heads. Its constants $0.08$, $0.20$, and $80$ were fixed before comparing models. Movie identity indexes training popularity. Genres are reserved for prediction-head features. Each seed samples three slate views per training user and two per validation or test user for model evaluation.

\paragraph{Grid-based routing score, distinct from candidate selection.}
The saved experiments use \texttt{grid\_predictions} to compute the routing score after a slate is selected. Let $\ell_i$ be the training-popularity logit of item $i$, let $\sigma$ be the logistic function, and use $16$ equally spaced shifts $\theta_g\in[-3,3]$. With observed prefix labels $y_i$, the discrete posterior is
\[
w_g\ \propto\ \exp\{-\theta_g^2/(2\cdot1.6^2)\}
\prod_{i\in\mathrm{obs}}\sigma(\ell_i+\theta_g)^{y_i}
[1-\sigma(\ell_i+\theta_g)]^{1-y_i}.
\]
For hidden-item component means $\mu_g=n^{-1}\sum_{i\in\mathrm{hidden}}\sigma(\ell_i+\theta_g)$, define $\bar\mu=\sum_gw_g\mu_g$ and $\widehat v=\sum_gw_g(\mu_g-\bar\mu)^2$. The empirical score is $S=n\widehat v/(|\bar\mu-\lceil cn\rceil/n|+.03)^2$. The $.03$ offset regularizes the boundary distance. Its value is an empirical regularizer separate from Proposition~\ref{prop:taxscore}. No neural-head output, hidden rating, tail label, or realized test gain enters $S$. This common grid surrogate can be misspecified. Candidate-rule perturbations share this fixed grid estimator.

\paragraph{Slices versus routing policies.}
The code uses test-score terciles for low/mid/high descriptive bins and test-score quantiles for top-10/20/30\% slices. Figure~\ref{fig:main-diagnostics} uses score quartiles. These ranks use test covariates and revealed prefixes, never hidden outcomes. The descriptive cutoffs consequently depend on the unlabeled test-score distribution. The original deployment switch instead obtains its numeric cutoff from a validation-score quantile and applies it unchanged to test scores, so the realized test fraction can differ from the nominal budget. The additional matched-budget study ranks each split separately to select exactly the requested number of slates. All fractions and algorithms are specified before outcome evaluation. Confidence thresholds and model choices use validation labels.

\paragraph{Prediction heads and payoff.}
GlobalGrid integrates conditional Poisson-binomial tail probabilities over the shared shift. SeqTXGlobal and ObjSetGlobal learn grid weights. ANP uses a diagonal Gaussian latent vector and FlowLatent a conditional affine-coupling flow. TwoTowerIndep completes item predictions independently. TwoTowerDropout averages Monte Carlo (MC) dropout tails. TwoTowerEnsMean averages member item probabilities before aggregation, whereas TwoTowerEnsMix averages member tail probabilities. SeqTXCountPM and ObjSetTFPM increase direct-count capacity without introducing a shared latent. For a predicted event probability $p$, the implemented policy predicts $\mathbf1[p\ge.5]$ if $\max(p,1-p)\ge\tau$, otherwise defers. Its payoff is $1$ for a correct prediction, $0$ for an error, and $1-\lambda$ for deferral, with $\lambda=.30$ in the headline results and $\tau$ selected on validation users. It measures the payoff of this offline triage decision under a declared cost.

\paragraph{Matched-backbone and fixed-ensemble controls.}
The additional ML1M-80 study uses five seeds, a common 16-point scoring grid, and 64 SeqTXGlobal components. The Transformer backbone is unchanged. The count output has no sampling parameter, so matching 64 latent components to ANP/Flow's 64 samples matches posterior-evaluation resolution. Wall-clock cost and parameter count can differ. SeqTXGlobal has 240,737 parameters versus 145,377 for SeqTXCount. The separate parameter-matched controls address capacity. EnsMean and EnsMix reuse exactly the same three trained members (72,579 total parameters), with no extra training or member forward passes. The aggregation operation, including its tail calculations, differs. Tables~\ref{tab:matched-attribution} and~\ref{tab:ensemble-attribution} report the mean and population standard deviation of paired differences across the five training seeds.

\paragraph{Equal-budget routing controls.}
For each rich head, TaxScore, SeqTXCount uncertainty $p(1-p)$, and random routing select exactly 330 or 496 of 1,652 test slates. Random results average 100 masks per seed. Each selector/head/budget combination retunes its mixed-policy confidence threshold on validation groups. Seed-level means are then aggregated. TaxScore's incremental advantage over each comparator is positive in all five seeds for full NLL and utility at both budgets (Table~\ref{tab:matched-routing}). Uncertainty routing requires evaluating the cheap head first. Compute accounting includes this selector-specific work.

\paragraph{Deferral-cost sensitivity.}\label{app:cost-sensitivity}
The primary audit at $\lambda=.30$ reports ANP/Flow high-bin utility gains of $.027/.023$, and original validation-cutoff switch gains of $.008/.009$ at 20\% and $.010/.012$ at 30\%. A separate archived five-seed sensitivity summary retunes $\tau$ at each $\lambda\in\{.10,.20,.30,.40,.50\}$ with a fresh validation-selected operating point. Averaged across ANP and Flow, its high-slice gains over $.20$--$.50$ range from $.009$ to $.044$, with 20\%-switch gains from $.002$ to $.011$. At $.10$, differences are within $3\times10^{-4}$ of zero. These ranges come from the recorded sensitivity summary, which uses a separate aggregation from the headline results. The released archive retains that summary. Per-cost prediction arrays are unavailable, limiting reproducibility to the archived ranges. The matched-routing results retain their separate raw records and aggregation. With cheap deferral, improved posteriors may leave both the selected action and the realized downstream payoff unchanged.

\paragraph{Inference timing and routing overhead.}
Architecture-matched SeqTXCount, ANP, and Flow heads were timed on two NVIDIA L20 GPUs with preloaded $(82,23)$ inputs, float32 precision, TensorFloat-32 enabled, and 64 latent samples for ANP/Flow. Each GPU ran five trials: batch size 1 used 50 warm-ups and 200 timed batches per trial. Batch size 192 used 20 and 50. CUDA events with synchronization measure head output only. We report the range of per-GPU means across two clean runs. An overlapping-load run was excluded. Table~\ref{tab:latency-audit} also reports the original five-seed training times.

For direct routing of fraction $r$, the head-only estimate is $(1-r)L_{\rm cheap}+rL_{\rm rich}$. Including score construction and routing overhead $L_S$, savings relative to rich-only inference require $L_S<(1-r)(L_{\rm rich}-L_{\rm cheap})$. The timing protocol excludes the score's grid computation. End-to-end evaluation must measure this additional overhead. At batch size 192, interpolation further assumes each head receives full batches. Small queues may erase the expected throughput savings. Cross-domain use requires held-out validation of candidate rules and score regularization against utility and total cost.

\section{Supplementary Tables}
\label{app:tables}

\begin{paperblock}
\begin{minipage}{\linewidth}
\captionof{table}{Notation convention used throughout the paper.}
\label{tab:notation}
\small
\begin{tabular}{p{0.24\linewidth}p{0.66\linewidth}}
\toprule
Symbol & Meaning \\
\midrule
$R$ or $R_n$ & exact posterior count under the object-space/shared-latent law \\
$\widehat R$ or $\widehat R_n$ & approximate posterior count used for decisions \\
$M_n$ and $\widehat M_n$ & exact and approximate hidden empirical measures in the non-binary law \\
$\widetilde X_i$ or $\widehat Y_i$ & generated independent item under an independent completion \\
$G$ and $\bar G_O$ & directing random measure and its posterior mean after observing $O$ \\
$Q$ & scalar de Finetti latent frequency \\
$\Theta$ & general shared latent state, possibly high-dimensional \\
$W_i$ & observed covariates/metadata of hidden item $i$ \\
$q_i(\Theta)$ & conditional success probability of item $i$ given the shared latent \\
TaxScore & margin-normalized ambiguity diagnostic for a threshold decision \\
$S^+$ and $S^-$ & positive and negative signed covariance mass in Proposition~\ref{prop:signedcov} \\
\bottomrule
\end{tabular}
\end{minipage}
\end{paperblock}

\begin{paperblock}
\begin{minipage}{\linewidth}
\captionof{table}{Predictive representations of empirical baselines. Each role describes the implemented prediction head and the posterior information exposed to downstream decisions.}
\label{tab:coverage}
\setlength{\tabcolsep}{3pt}
\scriptsize
\begin{tabular}{p{0.20\linewidth}p{0.24\linewidth}p{0.48\linewidth}}
\toprule
Model & Interface role & Interpretation \\
\midrule
\multicolumn{3}{l}{\emph{Tax-risk interfaces}}\\
IndepPB & Theorem~\ref{thm:universal} & Mean-preserving independent completion. \\
MeanIndep & Prop.~\ref{prop:covlatent} & Heterogeneous independent marginal completion. \\
TwoTowerIndep & Prop.~\ref{prop:covlatent} & Item marginals are completed independently across the slate. \\
TwoTowerEnsMean & Tax-risk interface & Ensemble item marginals are averaged before tail computation. \\
\addlinespace[0.15em]
\multicolumn{3}{l}{\emph{Conditional finite-probe interfaces}}\\
SeqMLP & Conditional & Covered only when its posterior head is a fixed $\Psi$-probe compression. \\
SeqPermAvg & Conditional & Averaging fixed probes retains the same posterior information. \\
ARRollout & Conditional & Covered when every rollout uses the same fixed probe vector. \\
\addlinespace[0.15em]
\multicolumn{3}{l}{\emph{Stress tests and capacity controls}}\\
SeqGRUCount & Stress test & Direct count head beyond the exact lower-bound class. \\
SeqTXCount & Stress test & Permutation-averaged Transformer count head beyond the exact lower-bound class. \\
SeqTXCountPM & Capacity control & Parameter-matched direct count head. \\
ObjSetTFPM & Capacity control & Parameter-matched direct set/count head. No added shared latent. \\
\addlinespace[0.15em]
\multicolumn{3}{l}{\emph{Shared-sample and resolution mechanisms}}\\
TwoTowerDropout & Shared-sample baseline & MC dropout carries a shared stochastic state during slate scoring. \\
TwoTowerEnsMix & Shared-sample baseline & Slate-level tails are mixed over shared ensemble members. \\
GlobalGrid & Resolution design & Dimension-growing grid posterior for the shared latent. \\
SeqTXGlobal & Resolution design & Transformer backbone with an explicit global-latent posterior head. \\
ANP & Resolution design & Attentive neural-process posterior over a shared latent vector. \\
FlowLatent & Resolution design & Conditional flow posterior over a shared latent vector. \\
ObjMLP & Object-space comparator & Object-space posterior surrogate used as a competing design. \\
ObjSetTF & Object-space comparator & Object-space Set Transformer surrogate used as a stronger design. \\
ObjSetGlobal & Resolution design & Set backbone with an explicit global-latent posterior head. \\
\bottomrule
\end{tabular}
\end{minipage}
\end{paperblock}

\begin{paperblock}
\begin{minipage}{0.985\linewidth}
\centering
\captionof{table}{MovieLens tail-event NLL on full-distribution and high-TaxScore slices. Means and standard deviations are over five seeds. The high slice is the pre-specified decision-critical TaxScore bin. Dashes indicate baselines absent from the MovieLens 10M scale-up experiments.}
\label{tab:movie-full-high}
\scriptsize
\setlength{\tabcolsep}{3.2pt}
\begin{tabular}{llrrrrrr}
\toprule
Config & Slice & SeqTX & SeqTX-PM & 2T-Mean & 2T-Mix & ANP & Flow \\
\midrule
ML1M-40 & All & 0.639 $\pm$ 0.010 & 0.641 $\pm$ 0.009 & 0.696 $\pm$ 0.030 & 0.694 $\pm$ 0.029 & 0.625 $\pm$ 0.011 & 0.628 $\pm$ 0.009 \\
ML1M-40 & High & 0.662 $\pm$ 0.013 & 0.666 $\pm$ 0.009 & 0.678 $\pm$ 0.022 & 0.677 $\pm$ 0.021 & 0.637 $\pm$ 0.002 & 0.638 $\pm$ 0.006 \\
\addlinespace
ML1M-80 & All & 0.608 $\pm$ 0.009 & 0.607 $\pm$ 0.011 & 0.900 $\pm$ 0.085 & 0.890 $\pm$ 0.082 & 0.587 $\pm$ 0.012 & 0.589 $\pm$ 0.010 \\
ML1M-80 & High & 0.673 $\pm$ 0.011 & 0.665 $\pm$ 0.008 & 0.772 $\pm$ 0.056 & 0.767 $\pm$ 0.053 & 0.629 $\pm$ 0.008 & 0.628 $\pm$ 0.011 \\
\addlinespace
ML10M-120 & All & 0.618 $\pm$ 0.007 & 0.619 $\pm$ 0.006 & -- & -- & 0.594 $\pm$ 0.003 & 0.594 $\pm$ 0.006 \\
ML10M-120 & High & 0.670 $\pm$ 0.007 & 0.672 $\pm$ 0.007 & -- & -- & 0.639 $\pm$ 0.012 & 0.639 $\pm$ 0.014 \\
\addlinespace
\bottomrule
\end{tabular}
\end{minipage}
\end{paperblock}

\begin{paperblock}
\begin{minipage}{0.985\linewidth}
\centering
\captionof{table}{Pre-specified high-TaxScore triage slices on MovieLens. The top slices are selected by the TaxScore diagnostic computed from visible metadata and observed prefix ratings.}
\label{tab:movie-tail-triage}
\scriptsize
\setlength{\tabcolsep}{3.0pt}
\begin{tabular}{llrrrrrrr}
\toprule
Config & Slice & Count & SeqTX NLL & SeqTX util. & ANP NLL & ANP util. & Flow NLL & Flow util. \\
\midrule
ML1M-80 & Top 20\% & 331 & 0.686 $\pm$ 0.008 & 0.688 $\pm$ 0.014 & 0.631 $\pm$ 0.011 & 0.722 $\pm$ 0.007 & 0.628 $\pm$ 0.011 & 0.720 $\pm$ 0.010 \\
ML1M-80 & Top 10\% & 166 & 0.690 $\pm$ 0.015 & 0.689 $\pm$ 0.022 & 0.609 $\pm$ 0.013 & 0.735 $\pm$ 0.018 & 0.603 $\pm$ 0.016 & 0.729 $\pm$ 0.009 \\
\addlinespace
ML10M-120 & Top 20\% & 500 & 0.675 $\pm$ 0.009 & 0.683 $\pm$ 0.011 & 0.639 $\pm$ 0.011 & 0.715 $\pm$ 0.006 & 0.638 $\pm$ 0.008 & 0.712 $\pm$ 0.013 \\
ML10M-120 & Top 10\% & 250 & 0.675 $\pm$ 0.020 & 0.680 $\pm$ 0.015 & 0.642 $\pm$ 0.010 & 0.712 $\pm$ 0.004 & 0.645 $\pm$ 0.006 & 0.709 $\pm$ 0.010 \\
\addlinespace
\bottomrule
\end{tabular}
\end{minipage}
\end{paperblock}

\begin{paperblock}
\begin{minipage}{0.985\linewidth}
\centering
\captionof{table}{Original budgeted switch gains over SeqTXCount. Mixed predictions use the rich head on slates exceeding a validation-score cutoff and the cheap head elsewhere. Full gains average all test slates. Flagged gains average the routed subset. The offline experiment evaluates both heads to form these comparisons. Latency is assessed separately from this payoff calculation. See Table~\ref{tab:matched-routing} for the separate exact-budget comparison and Table~\ref{tab:latency-audit} for head-only timings.}
\label{tab:movie-deployment-switch}
\scriptsize
\setlength{\tabcolsep}{3.0pt}
\begin{tabular}{lllrrrr}
\toprule
Config & Budget & Target & Test frac. & Full NLL gain & Full util. gain & Per-flagged util. gain \\
\midrule
ML1M-80 & 10\% & ANP & 10.0\% & 0.008 $\pm$ 0.001 & 0.004 $\pm$ 0.004 & 0.045 $\pm$ 0.031 \\
ML1M-80 & 20\% & ANP & 19.6\% & 0.011 $\pm$ 0.001 & 0.008 $\pm$ 0.003 & 0.035 $\pm$ 0.018 \\
ML1M-80 & 30\% & ANP & 29.1\% & 0.014 $\pm$ 0.002 & 0.010 $\pm$ 0.002 & 0.030 $\pm$ 0.010 \\
\addlinespace
ML1M-80 & 10\% & Flow & 10.0\% & 0.009 $\pm$ 0.001 & 0.004 $\pm$ 0.004 & 0.040 $\pm$ 0.023 \\
ML1M-80 & 20\% & Flow & 19.6\% & 0.012 $\pm$ 0.001 & 0.009 $\pm$ 0.003 & 0.032 $\pm$ 0.012 \\
ML1M-80 & 30\% & Flow & 29.1\% & 0.015 $\pm$ 0.003 & 0.012 $\pm$ 0.003 & 0.026 $\pm$ 0.008 \\
\addlinespace
ML10M-120 & 10\% & ANP & 9.7\% & 0.003 $\pm$ 0.002 & 0.003 $\pm$ 0.003 & 0.034 $\pm$ 0.020 \\
ML10M-120 & 20\% & ANP & 19.4\% & 0.007 $\pm$ 0.004 & 0.006 $\pm$ 0.004 & 0.033 $\pm$ 0.013 \\
ML10M-120 & 30\% & ANP & 29.3\% & 0.009 $\pm$ 0.006 & 0.006 $\pm$ 0.005 & 0.024 $\pm$ 0.012 \\
\addlinespace
ML10M-120 & 10\% & Flow & 9.7\% & 0.003 $\pm$ 0.002 & 0.004 $\pm$ 0.002 & 0.030 $\pm$ 0.020 \\
ML10M-120 & 20\% & Flow & 19.4\% & 0.008 $\pm$ 0.003 & 0.007 $\pm$ 0.004 & 0.029 $\pm$ 0.018 \\
ML10M-120 & 30\% & Flow & 29.3\% & 0.009 $\pm$ 0.005 & 0.007 $\pm$ 0.008 & 0.021 $\pm$ 0.017 \\
\addlinespace
\bottomrule
\end{tabular}
\end{minipage}
\end{paperblock}

\begingroup\scriptsize\setlength{\tabcolsep}{3pt}
\setlength{\LTcapwidth}{\linewidth}
\begin{longtable}{lllrrr}
\caption{\normalsize Mean paired-seed gains over SeqTXCount on MovieLens. Positive NLL gain means lower NLL for the target model. Positive utility gain means higher utility. NLL wins count seeds with positive NLL gains out of five. These headline-sweep top-slice gains use the full five-seed audit aggregation. Robustness Tables~\ref{tab:movie-proxy-robustness} and~\ref{tab:movie-threshold-robustness} use separate Flow-only selector-sweep aggregations.}\label{tab:movie-paired-ci}\\
\toprule
Config & Slice & Target & NLL gain & Utility gain & NLL wins/5 \\
\midrule\endfirsthead
\multicolumn{6}{l}{Table~\thetable\ (continued)}\\
\toprule
Config & Slice & Target & NLL gain & Utility gain & NLL wins/5 \\
\midrule\endhead
\midrule\multicolumn{6}{r}{Continued on the next page}\\\endfoot
\bottomrule\endlastfoot

ML1M-40 & All & SeqTX-PM & -0.002 & -0.001 & 2/5 \\
ML1M-40 & All & SeqTXGlobal & 0.001 & -0.002 & 3/5 \\
ML1M-40 & All & ANP & 0.014 & 0.002 & 5/5 \\
ML1M-40 & All & Flow & 0.010 & 0.005 & 5/5 \\
ML1M-40 & High & SeqTX-PM & -0.004 & -0.003 & 1/5 \\
ML1M-40 & High & SeqTXGlobal & 0.006 & 0.001 & 3/5 \\
ML1M-40 & High & ANP & 0.025 & 0.013 & 5/5 \\
ML1M-40 & High & Flow & 0.024 & 0.012 & 4/5 \\
\addlinespace
ML1M-80 & All & SeqTX-PM & 0.001 & 0.005 & 3/5 \\
ML1M-80 & All & SeqTXGlobal & 0.002 & 0.002 & 3/5 \\
ML1M-80 & All & ANP & 0.022 & 0.015 & 5/5 \\
ML1M-80 & All & Flow & 0.019 & 0.012 & 5/5 \\
ML1M-80 & High & SeqTX-PM & 0.008 & 0.003 & 3/5 \\
ML1M-80 & High & SeqTXGlobal & 0.018 & 0.007 & 5/5 \\
ML1M-80 & High & ANP & 0.044 & 0.027 & 5/5 \\
ML1M-80 & High & Flow & 0.044 & 0.023 & 5/5 \\
ML1M-80 & Top20 & SeqTX-PM & 0.008 & 0.003 & 3/5 \\
ML1M-80 & Top20 & ANP & 0.055 & 0.035 & 5/5 \\
ML1M-80 & Top20 & Flow & 0.058 & 0.032 & 5/5 \\
ML1M-80 & Top10 & SeqTX-PM & 0.014 & 0.002 & 3/5 \\
ML1M-80 & Top10 & ANP & 0.081 & 0.046 & 5/5 \\
ML1M-80 & Top10 & Flow & 0.087 & 0.040 & 5/5 \\
\addlinespace
ML10M-120 & All & SeqTX-PM & -0.002 & -0.003 & 2/5 \\
ML10M-120 & All & ANP & 0.024 & 0.014 & 5/5 \\
ML10M-120 & All & Flow & 0.024 & 0.014 & 5/5 \\
ML10M-120 & High & SeqTX-PM & -0.002 & 0.004 & 2/5 \\
ML10M-120 & High & ANP & 0.030 & 0.025 & 5/5 \\
ML10M-120 & High & Flow & 0.031 & 0.022 & 5/5 \\
ML10M-120 & Top20 & SeqTX-PM & 0.001 & 0.007 & 2/5 \\
ML10M-120 & Top20 & ANP & 0.036 & 0.032 & 5/5 \\
ML10M-120 & Top20 & Flow & 0.038 & 0.029 & 5/5 \\
ML10M-120 & Top10 & SeqTX-PM & 0.003 & 0.010 & 2/5 \\
ML10M-120 & Top10 & ANP & 0.033 & 0.032 & 4/5 \\
ML10M-120 & Top10 & Flow & 0.031 & 0.029 & 4/5 \\
\addlinespace

\end{longtable}\endgroup
\begin{paperblock}
\begin{minipage}{0.985\linewidth}
\centering
\captionof{table}{MovieLens diagnostic checks. Signed covariances are reported on the high-TaxScore slice. Correlations compare TaxScore and the finite-ratio proxy $n^2/N_{\rm user}$ against the ANP NLL gains measured separately in the low-, medium-, and high-score diagnostic bins.}
\label{tab:movie-diagnostic-audit}
\scriptsize
\setlength{\tabcolsep}{3.0pt}
\begin{tabular}{lrrrrrr}
\toprule
Config & $S_+$ & $S_-$ & $S_+/n(n-1)$ & neg. pair frac. & corr(TaxScore, NLL gain) & corr($n^2/N$, NLL gain) \\
\midrule
ML1M-40 & 59.5 & 0.0 & 0.038 & 0.000 & 0.83 & 0.56 \\
ML1M-80 & 243.1 & 0.0 & 0.038 & 0.000 & 0.95 & 0.74 \\
ML10M-120 & 591.8 & 0.0 & 0.041 & 0.000 & 0.75 & 0.42 \\
\bottomrule
\end{tabular}
\end{minipage}
\end{paperblock}

\begin{paperblock}
\begin{minipage}{0.985\linewidth}
\centering
\captionof{table}{MovieLens candidate-selection robustness. Variants perturb candidate budget $B$, prior-margin cutoff $m$, and prefix-ambiguity weight $w$. The subsequent grid-based TaxScore is unchanged. Hidden ratings, hidden positives, compared-head predictions, and test labels never enter candidate selection or scoring. Gains compare FlowLatent with SeqTXCount. Top-slice gains assess enrichment. Bin correlation is supplementary and can be negative, especially on the 1,200-user ML10M subset. The sign of switch utility additionally depends on action changes.}
\label{tab:movie-proxy-robustness}
\scriptsize
\setlength{\tabcolsep}{3.0pt}
\begin{tabular}{llrrr}
\toprule
Setting & Variant & corr(bin, NLL gain) & Top-20 NLL/util. gain & Switch util. gain \\
\midrule
ML10M-120 (1200 users) & $B=160$ & 0.76 & 0.039/0.008 & -0.001 \\
ML10M-120 (1200 users) & $B=40$ & 0.92 & 0.064/0.019 & 0.003 \\
ML10M-120 (1200 users) & $m=0.05$ & 0.20 & 0.046/0.004 & -0.004 \\
ML10M-120 (1200 users) & $m=0.10$ & 0.93 & 0.054/0.025 & 0.003 \\
ML10M-120 (1200 users) & $w=0.10$ & -0.09 & 0.044/0.028 & -0.002 \\
ML10M-120 (1200 users) & $w=0.30$ & -0.09 & 0.044/0.028 & -0.001 \\
ML10M-120 (1200 users) & default & -0.09 & 0.044/0.028 & -0.001 \\
ML10M-120 (1200 users) & prefix only & 0.98 & 0.053/-0.013 & 0.007 \\
ML10M-120 (1200 users) & prior only & -0.89 & 0.010/0.006 & -0.001 \\
ML1M-80 & $B=160$ & 0.98 & 0.037/0.023 & 0.006 \\
ML1M-80 & $B=40$ & 0.95 & 0.035/0.019 & 0.006 \\
ML1M-80 & $m=0.05$ & 0.93 & 0.033/0.012 & 0.003 \\
ML1M-80 & $m=0.10$ & 0.90 & 0.027/0.021 & 0.002 \\
ML1M-80 & $w=0.10$ & 0.92 & 0.054/0.029 & 0.008 \\
ML1M-80 & $w=0.30$ & 0.91 & 0.054/0.027 & 0.009 \\
ML1M-80 & default & 0.91 & 0.054/0.027 & 0.009 \\
ML1M-80 & prefix only & 1.00 & 0.035/0.017 & 0.005 \\
ML1M-80 & prior only & -0.59 & 0.002/0.001 & 0.000 \\
\bottomrule
\end{tabular}
\end{minipage}
\end{paperblock}

\begin{paperblock}
\begin{minipage}{0.985\linewidth}
\centering
\captionof{table}{ML1M-80 threshold robustness. The selector and model pair are fixed while the tail threshold $c$ changes. Gains compare FlowLatent with SeqTXCount. Positive values favor FlowLatent.}
\label{tab:movie-threshold-robustness}
\scriptsize
\setlength{\tabcolsep}{4.0pt}
\begin{tabular}{lrrr}
\toprule
Threshold $c$ & corr(bin, NLL gain) & Top-20 NLL/util. gain & Switch util. gain \\
\midrule
0.60 & 0.98 & 0.022/0.004 & 0.004 \\
0.70 & 0.91 & 0.054/0.027 & 0.009 \\
0.75 & 0.99 & 0.026/-0.005 & -0.002 \\
\bottomrule
\end{tabular}
\end{minipage}
\end{paperblock}

\begin{paperblock}
\begin{minipage}{0.985\linewidth}
\centering
\captionof{table}{Capacity and compute audit for the MovieLens $(k,n,c)=(2,80,0.7)$ setting. Parameter-matched direct count heads yield smaller high-TaxScore tail-calibration gains than shared-latent posterior heads. Train sec/run is the rounded per-seed training wall-clock segment recorded by the experiment script, so small equalities should be read at that reporting granularity.}
\label{tab:movie-capacity-audit}
\scriptsize
\begin{tabular}{lrrrr}
\toprule
Model & Params & Train sec/run & Samples / ensemble members & High NLL \\
\midrule
SeqTX & 145,377 & 3.0 & 1 & 0.673 $\pm$ 0.011 \\
SeqTX-PM & 437,585 & 3.0 & 1 & 0.665 $\pm$ 0.008 \\
SeqTXGlobal & 236,081 & 12.1 & 16 & 0.654 $\pm$ 0.013 \\
2T-Mean & 72,579 & 5.1 & 3 & 0.772 $\pm$ 0.056 \\
2T-Mix & 72,579 & 5.1 & 3 & 0.767 $\pm$ 0.053 \\
ANP & 91,329 & 15.5 & 64 & 0.629 $\pm$ 0.008 \\
Flow & 227,153 & 16.0 & 64 & 0.628 $\pm$ 0.011 \\
\bottomrule
\end{tabular}
\end{minipage}
\end{paperblock}

\begin{paperblock}
\begin{minipage}{0.985\linewidth}
\centering
\captionof{table}{Semi-synthetic quality-control batch inspection. Real UCI Breast Cancer features form item pools, while a shared latent risk controls batch defect prevalence. This mechanism check reports NLL and event-TV, with smaller values indicating more accurate aggregate posteriors.}
\label{tab:batch-inspection}
\scriptsize
\setlength{\tabcolsep}{3.0pt}
\begin{tabular}{llrr}
\toprule
TaxScore bin & Interface & NLL & Event-TV \\
\midrule
Low & indep. oracle & 0.050 $\pm$ 0.042 & 0.009 $\pm$ 0.000 \\
Low & shared oracle & 0.043 $\pm$ 0.028 & 0.000 $\pm$ 0.000 \\
Low & object head & 0.063 $\pm$ 0.024 & 0.029 $\pm$ 0.014 \\
\addlinespace
Mid & indep. oracle & 0.708 $\pm$ 0.036 & 0.059 $\pm$ 0.000 \\
Mid & shared oracle & 0.227 $\pm$ 0.008 & 0.000 $\pm$ 0.000 \\
Mid & object head & 0.269 $\pm$ 0.035 & 0.060 $\pm$ 0.024 \\
\addlinespace
High & indep. oracle & 0.732 $\pm$ 0.074 & 0.187 $\pm$ 0.002 \\
High & shared oracle & 0.535 $\pm$ 0.028 & 0.000 $\pm$ 0.000 \\
High & object head & 0.623 $\pm$ 0.039 & 0.135 $\pm$ 0.011 \\
\addlinespace
\bottomrule
\end{tabular}
\end{minipage}
\end{paperblock}

% Generated by make_rebuttal_tables.py from verified archived measurements.

\begin{paperblock}
\begin{minipage}{0.985\linewidth}
\captionof{table}{Same-backbone ML1M-80 control: SeqTXGlobal-64 relative to SeqTXCount. Entries are mean $\pm$ population standard deviation of paired differences over five seeds. Positive NLL reduction and utility gain favor the latent head. The backbone is fixed, but the output heads differ in parameter count. 64 components match latent-evaluation resolution while runtime can differ.}
\label{tab:matched-attribution}
\centering\footnotesize
\setlength{\tabcolsep}{4pt}
\renewcommand{\arraystretch}{1.08}
\begin{tabular}{@{}lccc@{}}
\toprule
Population & NLL reduction & Utility gain & Positive seeds (NLL, util.) \\
\midrule
Full population & $-0.0032\pm0.0057$ & $0.0023\pm0.0034$ & 2/5, 4/5 \\
Top 20\% & $0.0228\pm0.0118$ & $0.0173\pm0.0134$ & 5/5, 5/5 \\
\bottomrule
\end{tabular}
\end{minipage}
\end{paperblock}

\begin{paperblock}
\begin{minipage}{0.985\linewidth}
\captionof{table}{Fixed-member averaging control: TwoTowerEnsMix relative to TwoTowerEnsMean on ML1M-80. The same three trained members (72,579 parameters) and predictions are reused. Only tail aggregation changes. NLL/Brier/ECE reductions are positive in all five seeds. Utility decreases slightly despite the consistently improved probability estimates on the same evaluation groups.}
\label{tab:ensemble-attribution}
\centering\footnotesize
\setlength{\tabcolsep}{4pt}
\renewcommand{\arraystretch}{1.08}
\begin{tabular}{@{}lcccc@{}}
\toprule
Population & NLL reduction & Brier reduction & ECE reduction & Utility change \\
\midrule
Full population & $0.0108\pm0.0034$ & $0.00066\pm0.00026$ & $0.00186\pm0.00105$ & $-0.0011\pm0.0030$ \\
Top 20\% & $0.0050\pm0.0022$ & $0.00071\pm0.00020$ & $0.00183\pm0.00046$ & $-0.0036\pm0.0054$ \\
\bottomrule
\end{tabular}
\end{minipage}
\end{paperblock}

\begin{paperblock}
\begin{minipage}{0.985\linewidth}
\captionof{table}{Equal-budget ML1M-80 routing: full-population improvements over SeqTXCount, with five-seed mean $\pm$ standard deviation. All selectors route exactly 330/1,652 or 496/1,652 test slates. Random rows average 100 masks within each seed. Cheap uncertainty is $p(1-p)$. Each mixed-policy confidence threshold is selected on validation users. TaxScore exceeds each comparator for both metrics in all five seeds at each head/budget setting. These estimates use exact budgets. The original switch instead transfers a validation cutoff with a variable achieved fraction.}
\label{tab:matched-routing}
\centering\footnotesize
\setlength{\tabcolsep}{4pt}
\renewcommand{\arraystretch}{1.08}
\begin{tabular}{@{}lllcc@{}}
\toprule
Rich head & Budget & Selector & NLL reduction & Utility gain \\
\midrule
ANP & 20\% & TaxScore & $0.0105\pm0.0009$ & $0.0071\pm0.0024$ \\
ANP & 20\% & Cheap uncertainty & $0.0058\pm0.0013$ & $0.0024\pm0.0024$ \\
ANP & 20\% & Random & $0.0045\pm0.0010$ & $0.0032\pm0.0011$ \\
ANP & 30\% & TaxScore & $0.0137\pm0.0018$ & $0.0089\pm0.0012$ \\
ANP & 30\% & Cheap uncertainty & $0.0085\pm0.0016$ & $0.0035\pm0.0016$ \\
ANP & 30\% & Random & $0.0066\pm0.0015$ & $0.0044\pm0.0016$ \\
FlowLatent & 20\% & TaxScore & $0.0106\pm0.0011$ & $0.0094\pm0.0026$ \\
FlowLatent & 20\% & Cheap uncertainty & $0.0050\pm0.0025$ & $0.0019\pm0.0027$ \\
FlowLatent & 20\% & Random & $0.0037\pm0.0007$ & $0.0030\pm0.0019$ \\
FlowLatent & 30\% & TaxScore & $0.0127\pm0.0019$ & $0.0098\pm0.0026$ \\
FlowLatent & 30\% & Cheap uncertainty & $0.0068\pm0.0032$ & $0.0035\pm0.0029$ \\
FlowLatent & 30\% & Random & $0.0058\pm0.0011$ & $0.0042\pm0.0024$ \\
\bottomrule
\end{tabular}
\end{minipage}
\end{paperblock}

\begin{paperblock}
\begin{minipage}{0.985\linewidth}
\captionof{table}{Compute audit. Training seconds are the original five-seed training segment (mean $\pm$ standard deviation). Inference ranges are the two per-GPU mean latencies on NVIDIA L20 with preloaded inputs, five timing trials per GPU, float32/TF32, and 64 posterior samples for ANP/Flow. Each inference entry is milliseconds per slate. These are architecture-matched, head-only measurements, excluding score construction, data transfer, and queueing, whose contributions require an end-to-end timing audit in the target production serving environment.}
\label{tab:latency-audit}
\centering\footnotesize
\setlength{\tabcolsep}{4pt}
\renewcommand{\arraystretch}{1.08}
\begin{tabular}{@{}lrccc@{}}
\toprule
Head & Parameters & Train seconds & Online ($B=1$) & Batched ($B=192$) \\
\midrule
SeqTXCount & 145,377 & $3.0\pm0.2$ & $0.58$--$0.70$ & $0.0043$--$0.0046$ \\
ANP & 91,329 & $15.5\pm1.8$ & $5.92$--$8.07$ & $0.0581$--$0.0595$ \\
FlowLatent & 227,153 & $16.0\pm1.3$ & $6.48$--$8.76$ & $0.0610$--$0.0627$ \\
\bottomrule
\end{tabular}
\end{minipage}
\end{paperblock}

\begin{paperblock}
\begin{minipage}{0.985\linewidth}
\captionof{table}{Weighted head-latency estimates derived from separate head measurements. The calculation is $(1-r)L_{\rm cheap}+rL_{\rm rich}$, in milliseconds per slate. Steady-state batched estimates assume each routed head can fill batches of 192. Deployment savings additionally depend on TaxScore computation, routing overhead, and the achieved per-head batching efficiency.}
\label{tab:routing-latency}
\centering\footnotesize
\setlength{\tabcolsep}{4pt}
\renewcommand{\arraystretch}{1.08}
\begin{tabular}{@{}lcccc@{}}
\toprule
Rich head & Fraction & Online ms/slate & Batched ms/slate & Online saving vs. rich-only \\
\midrule
ANP & 20\% & $1.64$--$2.17$ & $0.0151$--$0.0156$ & 72.2--73.1\% \\
FlowLatent & 20\% & $1.76$--$2.31$ & $0.0156$--$0.0162$ & 72.9--73.6\% \\
ANP & 30\% & $2.18$--$2.91$ & $0.0204$--$0.0211$ & 63.2--64.0\% \\
FlowLatent & 30\% & $2.35$--$3.11$ & $0.0213$--$0.0220$ & 63.8--64.4\% \\
\bottomrule
\end{tabular}
\end{minipage}
\end{paperblock}

\begin{paperblock}
\begin{minipage}{\linewidth}
\captionof{table}{Analytic majority witness for $m=9$, $k=3$, $q_0=0.2$, and $q_1=0.8$. Here $v_s=\Var(Q\mid S_3=s)$. The ambiguous rows $s=1,2$ exhibit the largest independent-completion error.}
\label{tab:synthetic}
\small
\begin{tabular}{ccccccc}
\toprule
$s$ & $\alpha_s$ & Exact tail $p_s$ & Independent tail $\widehat p_s$ & $|p_s-\widehat p_s|$ & $v_s$ & TV lb. \\
\midrule
0 & 0.0154 & 0.0117 & 0.0020 & 0.0097 & 0.0055 & 0.0055 \\
1 & 0.2000 & 0.1938 & 0.0875 & 0.1063 & 0.0576 & 0.0576 \\
2 & 0.8000 & 0.8062 & 0.9125 & 0.1063 & 0.0576 & 0.0576 \\
3 & 0.9846 & 0.9883 & 0.9980 & 0.0097 & 0.0055 & 0.0055 \\
\bottomrule
\end{tabular}
\end{minipage}
\end{paperblock}

\begin{paperblock}
\begin{minipage}{\linewidth}
\captionof{table}{Digits hierarchical stop/reveal benchmark over five seeds. Prefix metrics are evaluated at $t=1$. Stopping thresholds are tuned on validation, with tuned $\tau$ denoting the validation-selected stopping probability threshold. Utility is maximized and the remaining error/cost metrics minimized.}
\label{tab:digits}
\setlength{\tabcolsep}{3.5pt}
\footnotesize
\begin{tabular}{lccccccc}
\toprule
Model & Prefix acc. & Prefix NLL & Prefix Brier & Prefix ECE & Utility & Avg. reveals & Tuned $\tau$ \\
\midrule
IndepPB & \mstd{0.733}{0.011} & \mstd{2.073}{0.178} & \mstd{0.241}{0.013} & \mstd{0.230}{0.013} & \mstd{0.753}{0.013} & \mstd{1.279}{0.071} & \mstd{0.89}{0.02} \\
SeqMLP & \mstd{\textbf{0.769}}{0.014} & \mstd{0.558}{0.024} & \mstd{0.182}{0.009} & \mstd{0.073}{0.016} & \mstd{\textbf{0.774}}{0.015} & \mstd{\textbf{1.228}}{0.050} & \mstd{0.64}{0.02} \\
SeqPermAvg & \mstd{\textbf{0.769}}{0.014} & \mstd{0.558}{0.024} & \mstd{0.182}{0.009} & \mstd{0.073}{0.016} & \mstd{0.773}{0.012} & \mstd{1.589}{0.712} & \mstd{0.68}{0.09} \\
ObjMLP & \mstd{0.765}{0.014} & \mstd{\textbf{0.553}}{0.024} & \mstd{\textbf{0.180}}{0.009} & \mstd{\textbf{0.055}}{0.011} & \mstd{0.773}{0.018} & \mstd{1.307}{0.167} & \mstd{0.68}{0.05} \\
\bottomrule
\end{tabular}
\end{minipage}
\end{paperblock}

\begin{paperblock}
\begin{minipage}{\linewidth}
\captionof{table}{Auxiliary hierarchical bag-majority benchmarks over five seeds. Lower is better for NLL, Brier, and ECE. These tasks are intentionally less tail-sensitive than the threshold-risk scaling task.}
\label{tab:bags}
\setlength{\tabcolsep}{4pt}
\footnotesize
\begin{tabular}{llcccc}
\toprule
Dataset & Model & Acc. & NLL & Brier & ECE \\
\midrule
Breast & IndepPB & \mstd{0.803}{0.009} & \mstd{0.552}{0.046} & \mstd{0.116}{0.003} & \mstd{0.050}{0.007} \\
Breast & SeqMLP & \mstd{0.801}{0.011} & \mstd{0.419}{0.011} & \mstd{0.134}{0.004} & \mstd{\textbf{0.043}}{0.011} \\
Breast & SeqPermAvg & \mstd{0.802}{0.009} & \mstd{0.412}{0.011} & \mstd{0.132}{0.004} & \mstd{0.046}{0.009} \\
Breast & ObjMLP & \mstd{\textbf{0.804}}{0.012} & \mstd{\textbf{0.393}}{0.024} & \mstd{0.124}{0.007} & \mstd{0.044}{0.011} \\
\midrule
Covtype & IndepPB & \mstd{0.786}{0.003} & \mstd{0.466}{0.009} & \mstd{0.148}{0.003} & \mstd{0.040}{0.003} \\
Covtype & SeqMLP & \mstd{0.760}{0.008} & \mstd{0.515}{0.016} & \mstd{0.166}{0.004} & \mstd{0.045}{0.012} \\
Covtype & SeqPermAvg & \mstd{0.775}{0.005} & \mstd{0.469}{0.005} & \mstd{0.153}{0.002} & \mstd{0.024}{0.004} \\
Covtype & ObjMLP & \mstd{\textbf{0.788}}{0.008} & \mstd{\textbf{0.455}}{0.008} & \mstd{\textbf{0.147}}{0.003} & \mstd{\textbf{0.024}}{0.005} \\
\bottomrule
\end{tabular}
\end{minipage}
\end{paperblock}

\begin{paperblock}
\begin{minipage}{\linewidth}
\captionof{table}{Natural grouped Brown Creeper stop/reveal benchmark over five seeds. Prefix metrics are evaluated after one revealed segment. Point metrics use the validation-selected threshold and reveal cost from the $\{0.00,0.02,\dots,0.10\}$ sweep. Larger accuracy and utility indicate improvement. Smaller NLL and ECE indicate better probabilities, and fewer reveals reduce observation costs.}
\label{tab:birds}
\setlength{\tabcolsep}{4pt}
\footnotesize
\begin{tabular}{lccccc}
\toprule
Model & Prefix acc. & Prefix NLL & Prefix ECE & Utility & Avg. reveals \\
\midrule
SeqGRU & \mstd{0.727}{0.028} & \mstd{0.534}{0.044} & \mstd{0.098}{0.032} & \mstd{0.789}{0.066} & \mstd{\textbf{2.051}}{0.466} \\
SeqTX & \mstd{\textbf{0.740}}{0.052} & \mstd{\textbf{0.530}}{0.047} & \mstd{0.095}{0.023} & \mstd{0.785}{0.051} & \mstd{2.452}{0.682} \\
ObjMLP & \mstd{0.700}{0.022} & \mstd{0.532}{0.032} & \mstd{\textbf{0.089}}{0.013} & \mstd{0.796}{0.041} & \mstd{2.372}{0.570} \\
ObjSetTF & \mstd{0.733}{0.049} & \mstd{0.558}{0.036} & \mstd{0.104}{0.027} & \mstd{\textbf{0.796}}{0.043} & \mstd{2.290}{0.643} \\
\bottomrule
\end{tabular}
\end{minipage}
\end{paperblock}

\section{Supplementary Figures}
\label{app:figures}

\begin{paperblock}
\begin{minipage}{\linewidth}
\includegraphics[width=0.985\linewidth]{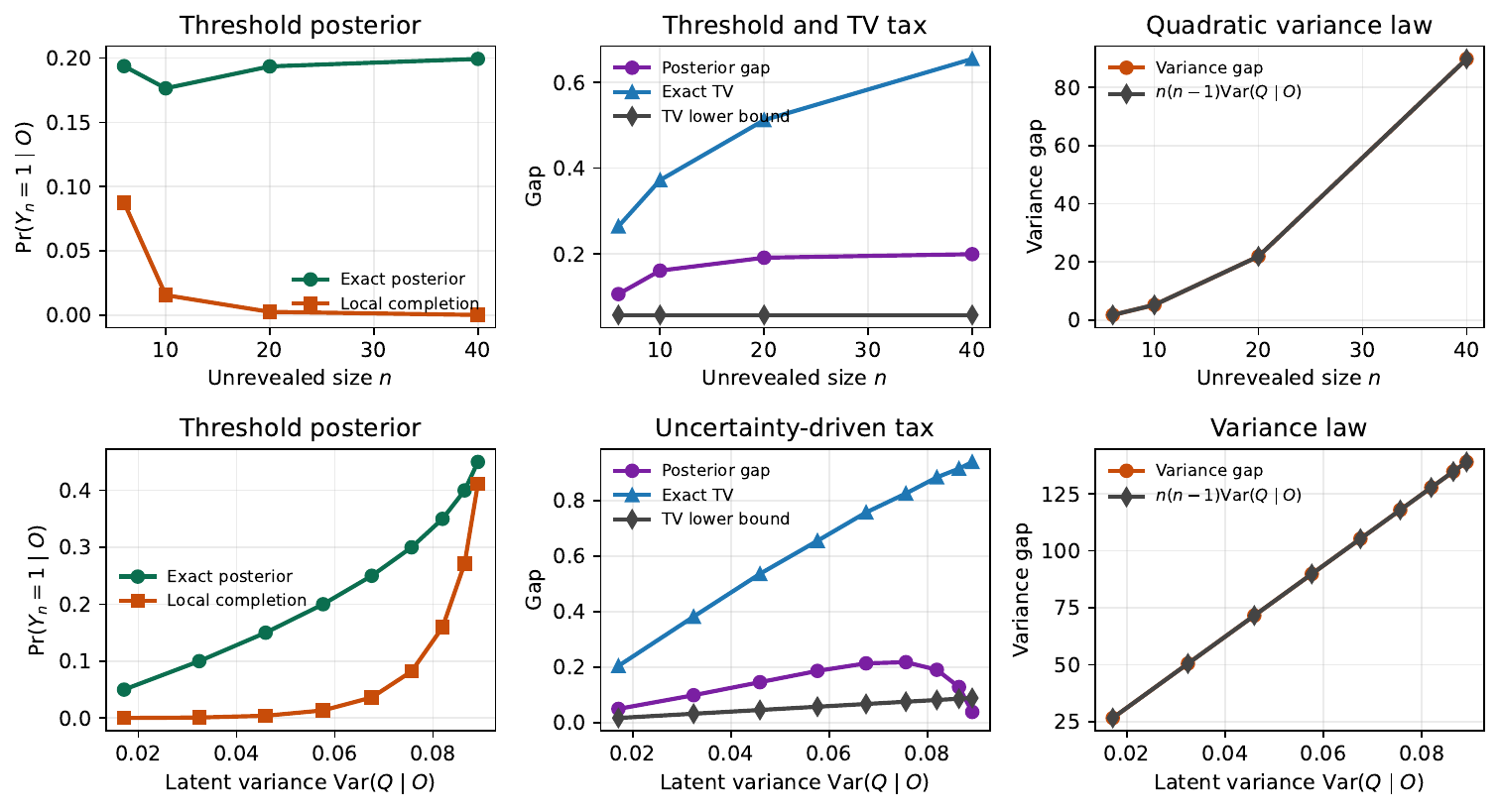}
\captionof{figure}{Synthetic law visualizations in the two-point family. As the unrevealed size $n$ or latent posterior variance increases, the threshold-posterior gap grows as predicted by Proposition~\ref{prop:taxscore} and Corollary~\ref{cor:thresholdrate}, while the TV and predictive-variance gaps follow Corollary~\ref{cor:tv}.}
\label{fig:scaling}
\end{minipage}
\end{paperblock}

\begin{paperblock}
\begin{minipage}{\linewidth}
\includegraphics[width=0.985\linewidth]{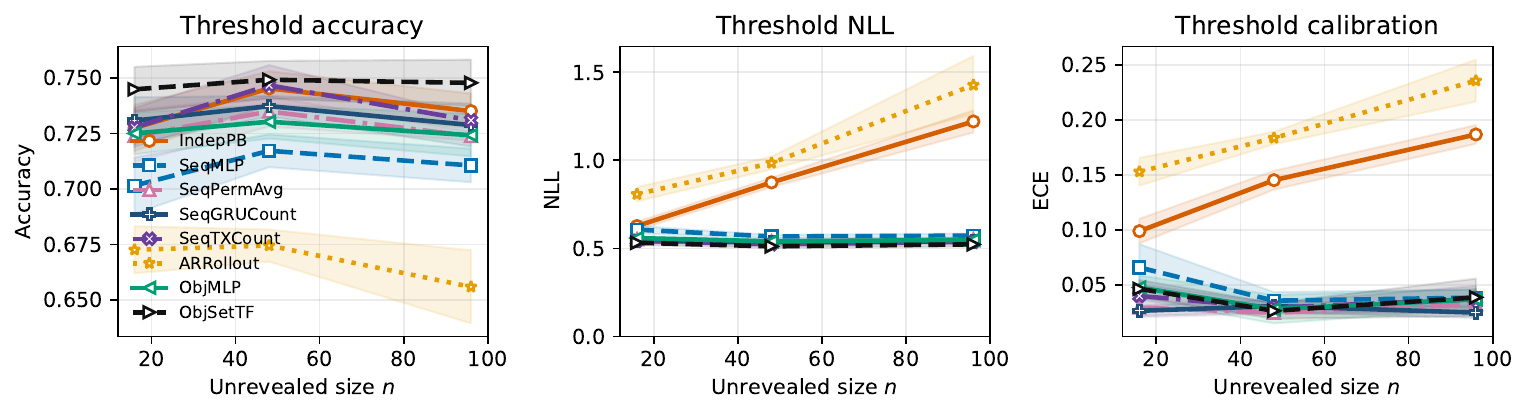}
\captionof{figure}{Covtype threshold-risk scaling over five seeds. Independent completion remains deceptively competitive on accuracy, but calibration and NLL deteriorate sharply with hidden size. GRU and permutation-averaged Transformer count heads improve on rollout. The object-space Set Transformer achieves the strongest calibration and NLL across the evaluated hidden sizes.}
\label{fig:threshold}
\end{minipage}
\end{paperblock}

\begin{paperblock}
\begin{minipage}{\linewidth}
\includegraphics[width=0.985\linewidth]{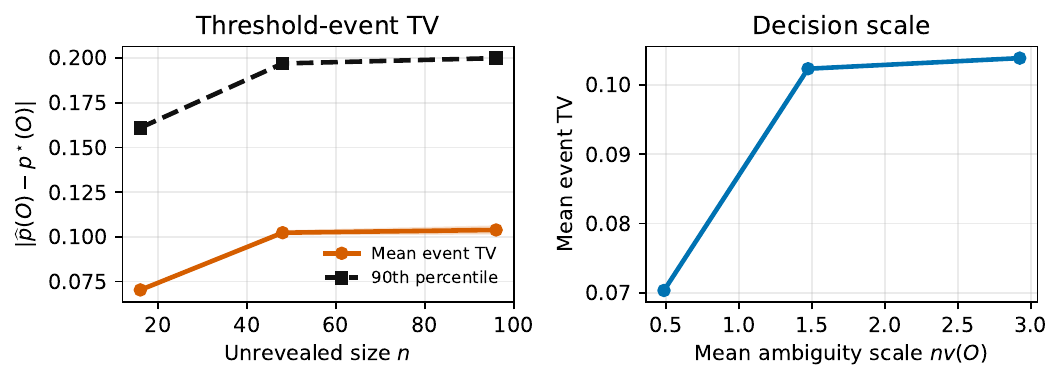}
\captionof{figure}{Direct theorem-aligned diagnostic on the covtype hierarchical threshold task. The exact oracle threshold posterior $p^\star(O)$ is compared with the oracle mean-preserving independent completion $\widehat p(O)$ using the binary-event TV $|\widehat p(O)-p^\star(O)|$, the maximal expectation error over binary-event functions with values in $[0,1]$ under the exact and independently completed predictive laws.}
\label{fig:oracle}
\end{minipage}
\end{paperblock}

\begin{paperblock}
\begin{minipage}{\linewidth}
\includegraphics[width=0.985\linewidth]{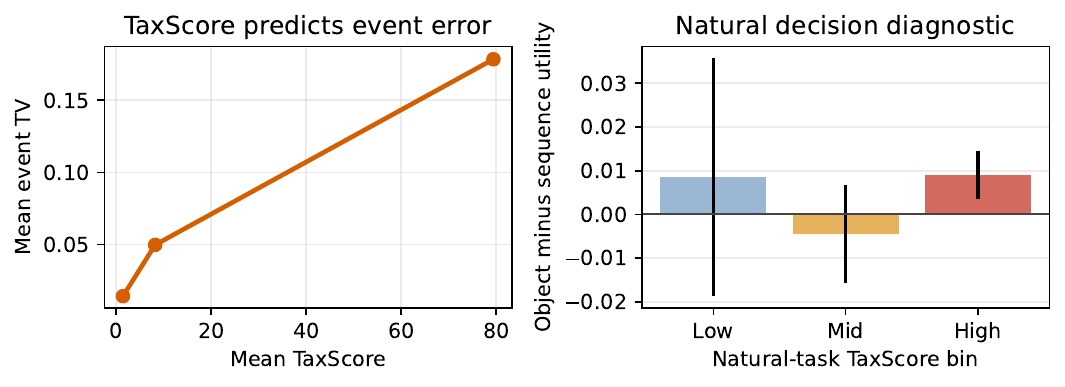}
\captionof{figure}{TaxScore diagnostics. Left: controlled covtype-style hierarchical bags binned by the margin-normalized ambiguity score from Proposition~\ref{prop:taxscore}. Event-TV increases sharply with TaxScore. Right: natural MIL and MovieLens slate triage tasks binned by an estimated TaxScore proxy. High-score bins show consistent object-space utility gains, with smaller gains in the lower-score regimes.}
\label{fig:taxscore}
\end{minipage}
\end{paperblock}

\begin{paperblock}
\begin{minipage}{\linewidth}
\includegraphics[width=0.55\linewidth]{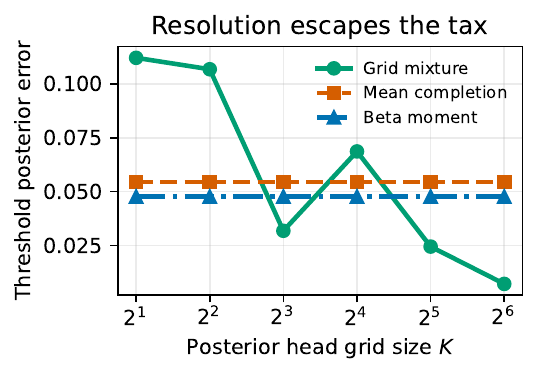}
\captionof{figure}{Posterior-head resolution ablation in the controlled Bernoulli mixture. A grid mixture head improves as its support resolution grows, while the fixed mean-completion and Beta moment summaries retain a threshold-probability error floor across all tested support resolutions.}
\label{fig:headres}
\end{minipage}
\end{paperblock}

\begin{paperblock}
\begin{minipage}{\linewidth}
\includegraphics[width=0.58\linewidth]{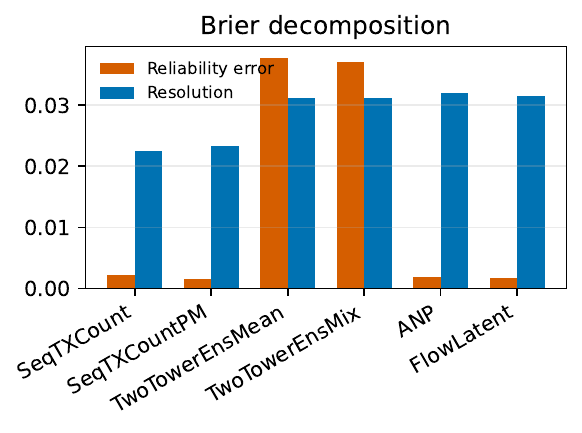}
\captionof{figure}{MovieLens Brier decomposition on the critical-slate audit. This decomposition complements the main-text reliability curve by separating its calibration and resolution components.}
\label{fig:brier}
\end{minipage}
\end{paperblock}

\begin{paperblock}
\begin{minipage}{0.49\linewidth}
\centering
\includegraphics[width=0.985\linewidth]{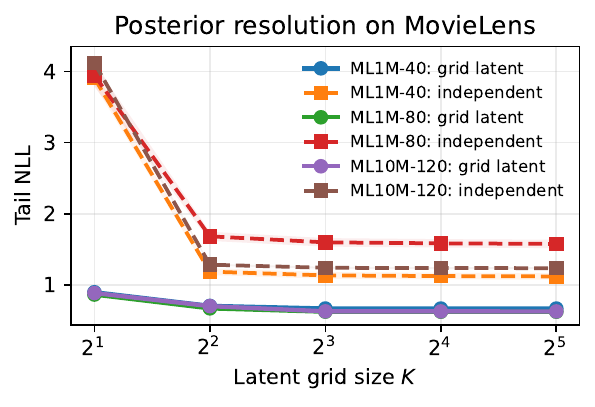}
\end{minipage}\hfill
\begin{minipage}{0.49\linewidth}
\centering
\includegraphics[width=0.985\linewidth]{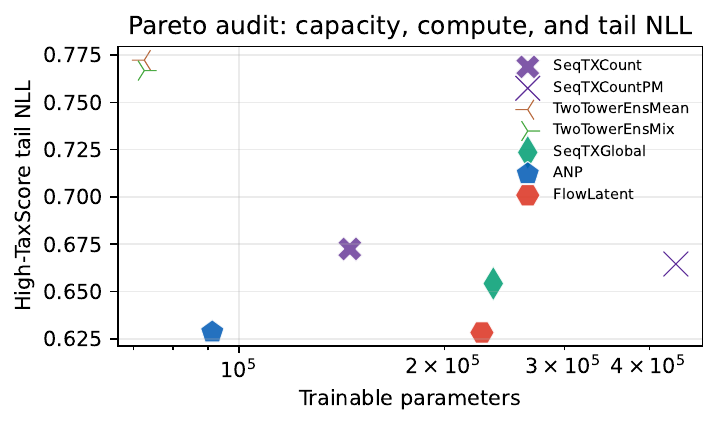}
\end{minipage}
\captionof{figure}{Resolution and capacity audits. Left: grid-latent posteriors improve tail NLL as resolution increases, while independent completions plateau. Right: parameter-matched direct count heads test whether additional capacity reproduces the gains from shared-latent prediction.}
\label{fig:main-resolution}
\end{paperblock}

\begin{paperblock}
\begin{minipage}{\linewidth}
\includegraphics[width=0.985\linewidth]{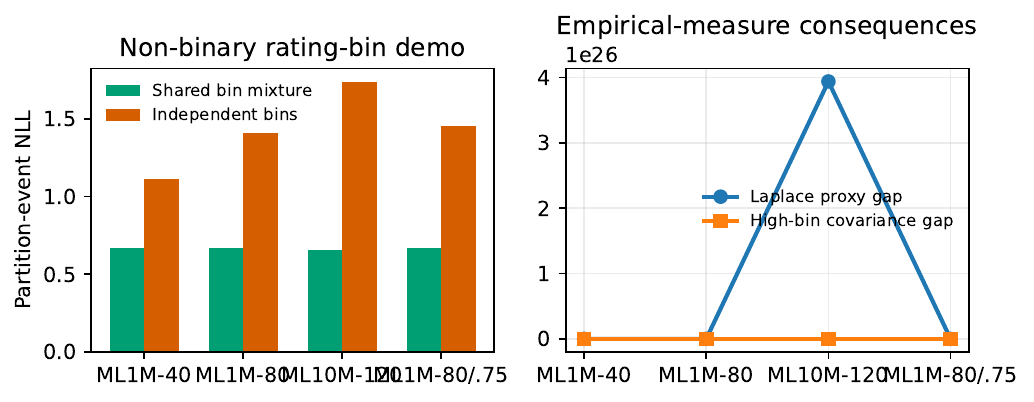}
\captionof{figure}{Non-binary MovieLens rating-bin demo for the empirical-measure law. Ratings are partitioned into low, middle, and high bins. The event depends jointly on high-bin and low-bin counts. The shared bin mixture improves partition-event NLL and exhibits the Laplace/covariance gaps predicted by Theorem~\ref{thm:partition}. The fixed Laplace-proxy scale emphasizes the large-slate gap. Smaller-slate differences are visually compressed by this common exponential scale.}
\label{fig:nonbinary}
\end{minipage}
\end{paperblock}

\begin{paperblock}
\begin{minipage}{\linewidth}
\includegraphics[width=0.62\linewidth]{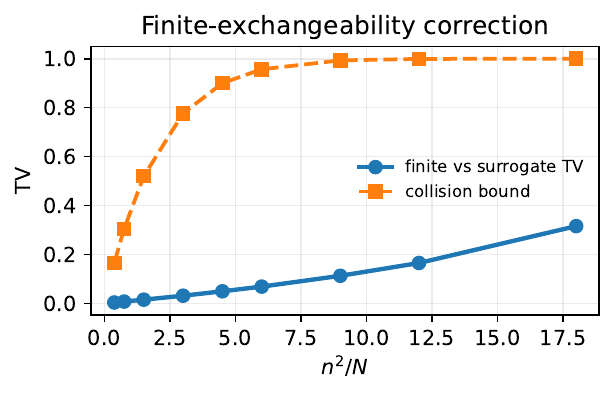}
\captionof{figure}{Finite-exchangeability stress test. A finite population with shared-latent composition is sampled without replacement and compared with the with-replacement de Finetti surrogate. The measured TV deviation decreases with $n^2/N$ within the collision correction in Corollary~\ref{cor:finitepop}.}
\label{fig:finite}
\end{minipage}
\end{paperblock}

\begin{paperblock}
\begin{minipage}{\linewidth}
\includegraphics[width=0.985\linewidth]{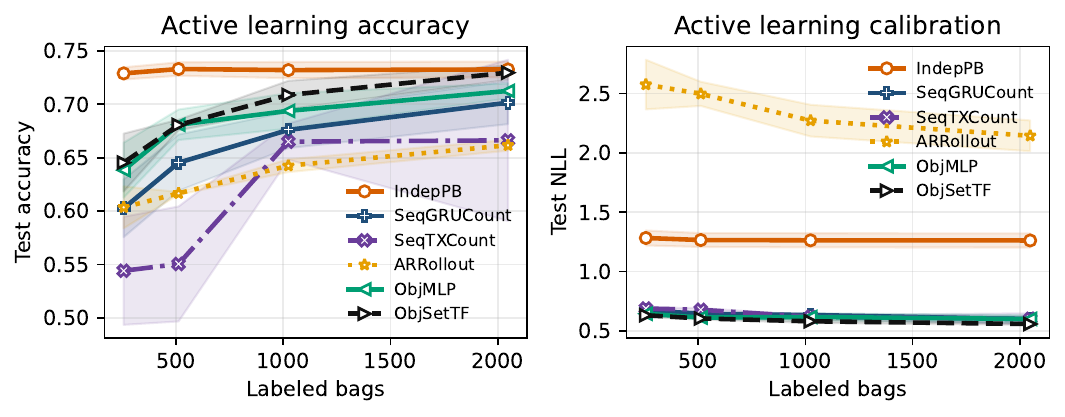}
\captionof{figure}{Covtype pool-based active learning over five seeds on the threshold-risk task with hidden size $n=96$. Each queried bag reveals its full item labels. Left: test accuracy versus labeled-bag budget. Right: test NLL versus labeled-bag budget. The object-space models are substantially more sample-efficient than rollout and remain competitive with stronger GRU/Transformer count-head baselines. Independent completion remains accuracy-competitive but essentially flat in NLL, highlighting the influence of posterior-family restrictions at larger labeling budgets.}
\label{fig:active}
\end{minipage}
\end{paperblock}

\begin{paperblock}
\begin{minipage}{\linewidth}
\includegraphics[width=0.985\linewidth]{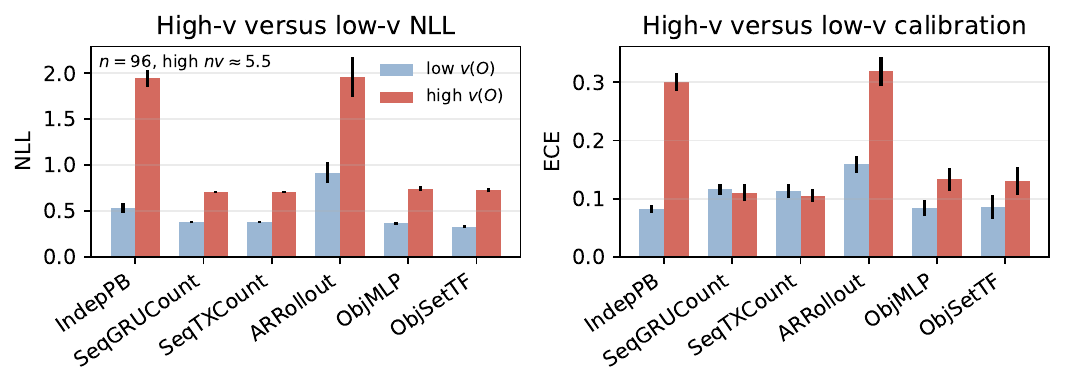}
\captionof{figure}{Covtype threshold-risk performance at the largest hidden size after splitting test prefixes by oracle latent posterior variance $v(O)$. The high-ambiguity slice has the largest independent-completion error, consistent with the hidden-size and residual-variance scale $n v(O)$ in Remark~\ref{rem:nv-scale}.}
\label{fig:ambiguity}
\end{minipage}
\end{paperblock}

\begin{paperblock}
\begin{minipage}{\linewidth}
\includegraphics[width=0.985\linewidth]{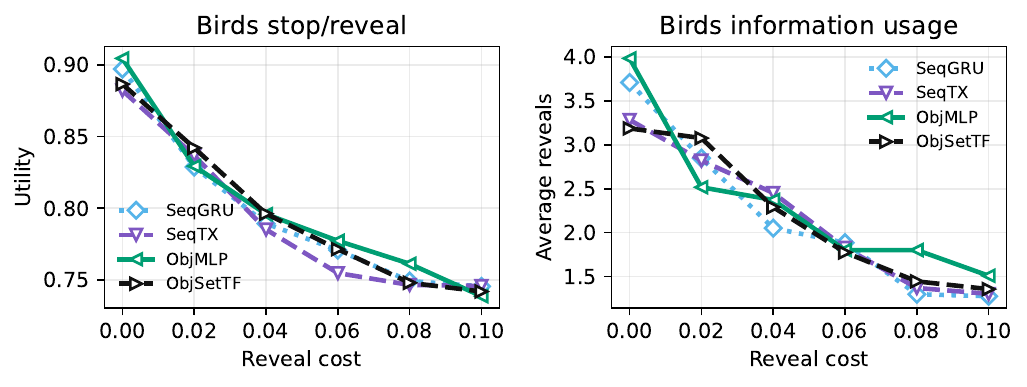}
\captionof{figure}{Natural grouped stop/reveal on the Brown Creeper multiple-instance benchmark over five seeds. Bags are audio recordings with local segment features and natural species-presence labels. No latent source is injected by the experiment. Left: utility as a function of reveal cost after re-tuning the stopping threshold for each cost. Right: the corresponding average number of reveals.}
\label{fig:birds}
\end{minipage}
\end{paperblock}

\begin{paperblock}
\begin{minipage}{\linewidth}
\includegraphics[width=0.985\linewidth]{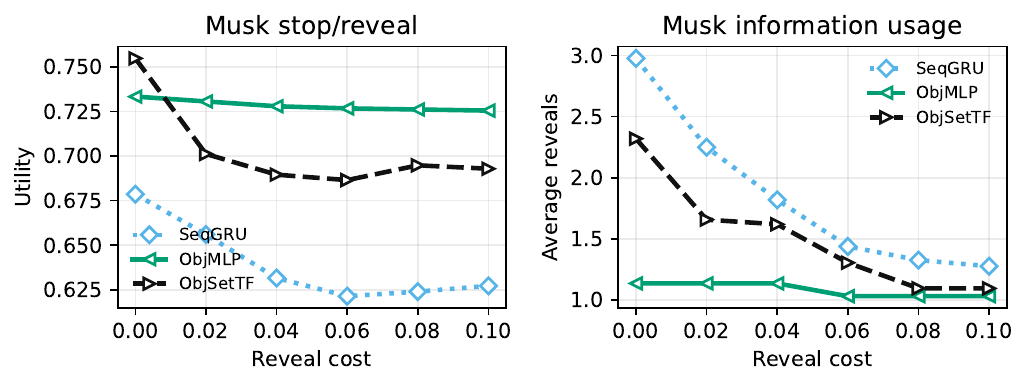}
\captionof{figure}{Natural grouped stop/reveal on the Musk multiple-instance dataset over five seeds. Each molecule is a bag of conformations. After each reveal the model may stop and classify the molecule or pay to reveal another conformation. Left: utility as a function of reveal cost after re-tuning the stopping threshold for each cost. Right: the corresponding average number of reveals.}
\label{fig:musk}
\end{minipage}
\end{paperblock}

\phantomsection
\section*{NeurIPS Paper Checklist}
\label{app:checklist}
\begin{enumerate}
\item \textbf{Claims and scope.}
\answerYes{} The paper analyzes independent marginalization of shared-latent predictive distributions and restricted finite-probe summaries. Classical convex-order ingredients are attributed explicitly. Separate posterior, threshold, and resolution results are distinguished from empirical routing claims. Theorem~\ref{thm:decision} bounds conditional Bayes regret from above. Positive regret additionally requires a change in the cost-dependent action.

\item \textbf{Limitations.}
\answerYes{} The limitations are stated where they matter technically. The Bernoulli convex-order theorem assumes a de Finetti/shared-latent posterior. Finite exchangeability is handled through explicit TV perturbation and finite-population correction in Appendix~A. The non-binary extension is a Laplace-functional law, and Appendix~B includes the counterexample explaining why the stronger statement is false. The finite-probe lower bound requires the specified information restriction on posterior summaries.

\item \textbf{Theory assumptions and proofs.}
\answerYes{} All formal assumptions are included in theorem or proposition statements. The setup defines the TV convention used in value-transfer constants. Appendix~A gives deferred consequences and Appendix~B gives proofs for the convex-order law, empirical-measure extension, heterogeneous-slate covariance tax, TaxScore bounds, finite-probe lower bound, resolution theorem, and finite-exchangeability robustness.

\item \textbf{Experimental protocol and reproducibility.}
\answerYes{} Section~\ref{sec:experiments} and Appendix~C specify data construction, user-level splits, allowed features, and scoring and routing rules. MovieLens candidate selection uses training popularity and revealed prefix ratings. The routing score uses a fixed grid posterior from the same visible information. Test-score ranks for descriptive slices are distinguished from validation cutoffs for the original switch and exact-budget routing in the added controls. Hidden test outcomes never determine selection.

\item \textbf{Code and data access.}
\answerYes{} The supplementary archive contains manuscript source, experiment implementations, available per-seed/aggregate JSON records, and reproduction scripts. The added attribution and latency tables are verified against raw saved records. The cost-sensitivity result has only an archived aggregate summary. That reproducibility limitation is disclosed in Appendix~C and the artifact notes. Datasets come from public scikit-learn, MIL, UCI, and MovieLens sources.

\item \textbf{Experimental details.}
\answerYes{} Dataset construction, hierarchical bag generation, MovieLens slate sampling, validation/test sizes, model-interface definitions, posterior-head choices, reveal-cost grids, and threshold-selection rules are specified in Appendix~C. The main text reports the headline settings. Appendix tables provide full-distribution, high-bin, top-quantile, calibration, capacity, signed-covariance, and deployment-switch audits.

\item \textbf{Error bars and statistical reporting.}
\answerYes{} Stochastic experiments use five seeds unless explicitly stated. Figures report mean curves or binned means. Appendix tables report mean $\pm$ standard deviation, paired mean gains, seed-consistency counts, and bin sizes where relevant. The MovieLens audit reports both full-distribution and protocol-selected slice metrics, separating critical-slate gains from average-population gains.

\item \textbf{Compute resources.}
\answerYes{} Appendix~C describes the Linux/NVIDIA L20 experiments and the separate two-GPU timing protocol. It distinguishes per-seed training segments, measured head latency, weighted routing estimates, and unmeasured scoring/transfer/queueing overhead.

\item \textbf{Ethics and broader impacts.}
\answerYes{} The paper is a theoretical and diagnostic study using public datasets. All experiments are offline analyses of existing public data. The intended impact is to improve calibration audits for tail-sensitive grouped decisions such as recommendation, quality control, active triage, and medical-panel-style risk assessment. We also state boundary cases where the diagnostic predicts little gain.

\item \textbf{Safeguards.}
\answerNA{} The supplementary material is for offline reproduction. It contains no private user data, no credentials, and no system for making real-world decisions. MovieLens user identifiers are public benchmark IDs and are used only within standard offline splits.

\item \textbf{Licenses for existing assets.}
\answerYes{} The paper cites the datasets and software sources used in experiments. Reproduction relies on public benchmark sources distributed through scikit-learn, MovieLens, UCI tabular datasets, and cited public MIL mirrors. No proprietary or restricted dataset is required to reproduce the reported results.

\item \textbf{New assets.}
\answerNo{} The released assets consist of code, aggregate outputs, and a pre-specified MovieLens audit protocol applied to existing datasets.

\item \textbf{Human subjects and crowdsourcing.}
\answerNA{} The work uses existing public benchmark data exclusively and involves no recruitment or direct interaction with human participants.
\end{enumerate}

\end{document}